\documentclass[10pt,twocolumn,letterpaper]{article}

\usepackage[review]{cvpr}      

\usepackage{pifont}

\usepackage[title]{appendix}
\usepackage{fontawesome}

\usepackage{amssymb}

\usepackage{multirow}
\usepackage{algorithm}
\usepackage{algpseudocode}
\usepackage{fixltx2e}
\usepackage{graphicx}
\usepackage{wrapfig,lipsum,booktabs}

\usepackage{arydshln}
\usepackage{array}
\usepackage[rightcaption]{sidecap}
\newcommand{\yes}{\textcolor{green}{\ding{51}}}  
\newcommand{\no}{\textcolor{red}{\ding{55}}}    
\usepackage{tabularx} 

\usepackage{threeparttable}
\usepackage[table,dvipsnames]{xcolor}
\usepackage{tikz}
\usepackage{adjustbox}
\usepackage{caption}
\definecolor{mygray}{gray}{.92}
\definecolor{customviolet}{RGB}{148, 0, 211}  

\makeatletter
\newcommand{\thickhline}{%
    \noalign {\ifnum 0=`}\fi \hrule height 1pt
    \futurelet \reserved@a \@xhline
}

\newcommand{\numbersBlue}[1]{\textcolor{blue}{\textbf{#1}}}

\definecolor{Green}{rgb}{0.85882353, 0.90980392, 0.84705882}

\definecolor{cvprblue}{rgb}{0.21,0.49,0.74}
\usepackage[pagebackref,breaklinks,colorlinks,citecolor=cvprblue]{hyperref}
\def\paperID{15161} 
\def\confName{CVPR}

\title{{\color{black} \textsc{DiffuseMix}}: Label-Preserving Data Augmentation with Diffusion Models}

\author{%
   Khawar Islam$^1$ \qquad Muhammad Zaigham Zaheer$^2$ 
   \qquad Arif Mahmood$^3$ \qquad
  Karthik Nandakumar$^2$ \\
  \small{$^1$\href{https://khawar-islam.github.io/}{FloppyDisk.AI} \qquad $^2$\href{https://mbzuai.ac.ae/}{Mohamed bin Zayed University of Artificial Intelligence}  \qquad $^3$\href{https://itu.edu.pk/}{Information Technology University, Punjab} } 
  \\
  \small{$^1${khawarr.islam@gmail.com} \qquad $^2$\{zaigham.zaheer, karthik.nandakumar\}@mbzuai.ac.ae} \qquad $^3${arif.mahmood@itu.edu.pk}  
}

\begin{document}

\twocolumn[{
\renewcommand\twocolumn[1][]{#1}
\maketitle
\begin{center}
    \raggedleft
    \vspace{-1.4em}
    \captionsetup{type=figure}
    \includegraphics[width=0.994\linewidth]{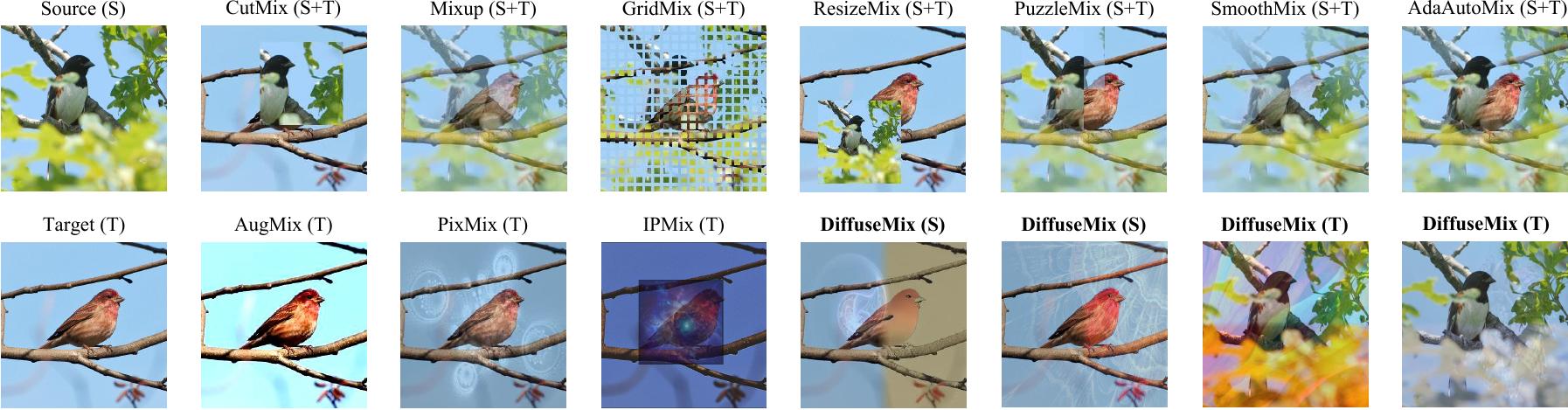}
    \vspace{-0.9em}
    \caption{\textbf{Top row:} existing mixup methods \emph{interpolate} two different training images \cite{mixup, kim2020co}. \textbf{Bottom row:} label-preserving methods. For each input image, \textsc{DiffuseMix} employs \emph{conditional prompts} to obtain generated images. The input image is then concatenated with a generated image to obtain a hybrid image. Each hybrid image is blended with a random fractal to obtain the final training image.
    }
    \label{fig:overview}
\end{center}}]

\renewcommand{\thefootnote}{\fnsymbol{footnote}}

\begin{abstract}


Recently, a number of image-mixing-based augmentation techniques have been introduced to improve the generalization of deep neural networks. In these techniques, two or more randomly selected natural images are mixed together to generate an augmented image. Such methods may not only omit important portions of the input images but also introduce label ambiguities by mixing images across labels resulting in misleading supervisory signals. 
To address these limitations, we propose \textsc{DiffuseMix}, a novel data augmentation technique that leverages a diffusion model to reshape training images,  supervised by our bespoke conditional prompts.
First, concatenation of a partial natural image and its generated counterpart is obtained which helps in avoiding the generation of unrealistic images or label ambiguities. 
Then, to enhance resilience against adversarial attacks and improves safety measures, 
a randomly selected structural pattern from a set of fractal images is blended into the concatenated image to form the final augmented image for training.
Our empirical results on seven different datasets reveal that  \textsc{DiffuseMix} achieves superior performance compared to existing state-of-the-art methods on tasks including general classification, fine-grained classification, fine-tuning, data scarcity, and adversarial robustness. Augmented datasets and codes are available here: \href{https://diffusemix.github.io/}{https://diffusemix.github.io/}

\end{abstract}



\section{Introduction}
\label{sec:intro}





In the era of deep learning, image-mixing-based data augmentation techniques stand out for their simplicity and effectiveness in addressing the generalization of learning models toward testing scenarios \cite{yun2019cutmix,kim2020puzzle,huang2021snapmix,kim2020co,uddin2020saliencymix,kang2023guidedmixup,zhao2022mixpro,chen2022transmix,kim2020puzzle, miao2022instance, choi2022tokenmixup}.
These techniques ingeniously mix randomly selected natural images and their respective labels from the training dataset using a number of \textit{mixing} combinations to synthesize new augmented images and labels. Such a process often implies linear interpolation of data, resulting in the generation of novel training images. These approaches have been proven to be effective in improving the performance of deep models \cite{cutmix, mixup, han2022yoco, qin2023adversarial, huang2021snapmix},.
\par

\begin{table*}
\setlength{\tabcolsep}{2.1pt}
\renewcommand{\arraystretch}{1.05}
\centering
\caption{Comparison of different image mixing techniques: most methods utilize natural images as source and target except \cite{verma2019manifold} using hidden state. \textsc{DiffuseMix} uses a \emph{generated} image produced by a diffusion model leveraging \emph{conditional prompts} and a fractal image for augmentation. } 
\vspace{-0.6em}
\label{table:compar_study}
\scalebox{0.85}{
\begin{tabular}{l|lcccccccccc}
\hline\thickhline
\rowcolor{mygray} 
&  Input & \cellcolor{mygray}\begin{tabular}[c]{@{}c@{}}Mixup\\ \cite{mixup} \end{tabular} & \cellcolor{mygray}\begin{tabular}[c]{@{}c@{}}ManifoldMixup\\ \cite{verma2019manifold} \end{tabular} 

& \cellcolor{mygray}\begin{tabular}[c]{@{}c@{}}CutMix\\ \cite{cutmix} \end{tabular} 

& \cellcolor{mygray}\begin{tabular}[c]{@{}c@{}}SaliencyMix\\ \cite{uddin2020saliencymix} \end{tabular} 

& \cellcolor{mygray}\begin{tabular}[c]{@{}c@{}}StyleMix\\ \cite{hong2021stylemix} \end{tabular} 

& \cellcolor{mygray}\begin{tabular}[c]{@{}c@{}}PuzzleMix\\ \cite{kim2020puzzle} \end{tabular} 

& \cellcolor{mygray}\begin{tabular}[c]{@{}c@{}}CoMixup\\ \cite{kim2020co} \end{tabular}  

& \cellcolor{mygray}\begin{tabular}[c]{@{}c@{}}PixMix\\ \cite{hendrycks2022pixmix} \end{tabular} 

& \cellcolor{mygray}\begin{tabular}[c]{@{}c@{}}GuidMixup\\ \cite{kang2023guidedmixup} \end{tabular} 

& \cellcolor{mygray}\begin{tabular}[c]{@{}c@{}} \textbf{ \textsc{DiffuseMix}}    \end{tabular}  \\
\hline
\multirow{6}{*}{\rotatebox{90}{ \emph{Components}}} & Source image & \yes  & \yes  & \yes  & \yes  & \yes  & \yes  & \yes  & \yes  & \yes & \yes 
\\
& Target image & \yes  & \no & \yes  & \yes  & \yes & \yes  & \yes  & \yes  & \yes  &  \no
\\
& Fractal image & \no & \no & \no & \no & \no & \no & \no & \yes & \no  & \yes 
\\
& Textual Prompts & \no & \no & \no & \no & \no & \no & \no & \no & \no & \yes 
\\
& Interpolation & \yes & \yes & \no  & \no & \yes & \no & \no & \yes & \yes & \yes 
\\
& 
Concatenation & \no & \no & \yes & \yes & \yes & \yes & \yes & \no  & \yes  & \yes 
\\
\hline

\multirow{5}{*}{\rotatebox{90}{ \emph{Tasks} }}
& Adversarial Robustness & \yes & \yes & \yes & \yes & \yes & \yes & \yes & \yes   & \no & \yes  
\\
& General Classification & \yes & \yes & \yes & \yes & \yes & \yes &  \yes &  \yes & \yes & \yes  
\\
& Fine Grained & \no  &  \no & \no & \no & \no & \no & \no  & \no & \yes & \yes  
\\
& Transfer Learning & \no &  \no & \yes & \yes  & \no & \no & \no & \no & \no & \yes  
\\
& Data Scarcity & \no  &  \no & \no & \no  & \no & \no & \no & \no & \yes & \yes 
\\

\hline
\end{tabular}
}
\vspace{-1em}
\end{table*}

However, these techniques may face a number of challenges such as the omission of salient image regions (Figure \ref{fig:overview}) and label ambiguities due to random placements of images \cite{kim2020puzzle}. A few researchers have attempted to alleviate these issues by introducing saliency-based mixup strategies in which important regions of one image are pasted onto the less important portions (mainly context) of another image \cite{kim2020puzzle, uddin2020saliencymix,kang2023guidedmixup}. These methods not only suffer from the costs but also the shortcomings of saliency detection methods. 
Moreover, as these methods still rely on mixing of two or more images belonging to different classes, the underlying issue of omitting the important context still persists. 





Recently, Diffusion Models \cite{takagi2023high, du2023stable, luo2023camdiff, saharia2022palette, dhariwal2021diffusion} have emerged as transformative approaches offering image-to-image generation and editing processes. Although the idea of using images generated by diffusion models directly as augmented images for training a classifier has been studied by some researchers \cite{trabucco2023effective, azizi2023synthetic}, this way of data augmentation does not result in significant performance gains. In fact, as reported in \cite{azizi2023synthetic}, the model trained using generated images directly as augmentation may even result in lower performance than the baseline trained without any augmentation. 
The underlying problem can be attributed to the limited control that these diffusion models offer over generated images. Owing to the sensitivity of diffusion models to conditional prompts, 
generation of desired complex \textcolor{customviolet}{\emph{scenes}, \emph{layouts}, and \emph{shapes}} in an image is a cumbersome task \cite{zhang2023adding}.
Thus, poorly constructed prompts pose the risk of producing images that may not be suitable for data augmentation as the generated images may deviate drastically from the actual data distribution. Training on such images may result in overfitting of the learning model on wrong data distribution, consequently resulting in performance degradation. 
Therefore, careful selection of prompts for data augmentation needs further investigation. Moreover, to mitigate the risk of poorly generated images affecting the overall training, a more efficient way of utilizing the generated images is necessary. 


To this end, we propose a novel data augmentation method, \textsc{DiffuseMix}, that leverages the capabilities of a Stable Diffusion model to generate diverse samples based on our tailored conditional prompts.
In contrast to Trabucco et al. \cite{trabucco2023effective}, rather than solely relying on Stable Diffusion for augmentation, we propose an effective approach which utilizes both original and generated images to create hybrid images. This way, visual diversity obtained by the diffusion models is \textit{infused} with the original images while retaining the key semantics.
In addition, to increase overall structural diversity, we blend self-similarity-fractals with the hybrid images to create the final training images. This blending 
has previously been found useful for ML safety measures \cite{hendrycks2022pixmix,huang2024ipmix} while in our approach, this added diversity 
helps in avoiding overfitting on the generated contents resulting in performance improvements. Our experimental results show that \textsc{DiffuseMix} benchmarks better generalization as well as increased adversarial robustness compared to the existing state-of-the-art (SOTA) augmentation methods. Moreover, it offers compatibility with a broad spectrum of datasets and can be incorporated into the training of various existing architectures. 
Some notable aspects of this research work are as follows:
\begin{itemize}
    \item We introduce a new data augmentation method driven by a diffusion model, which generates diverse images via our bespoke conditional prompts.
    
    \item We propose to \emph{concatenate a portion of the natural image with its generative counterpart}  to obtain hybrid images. The combination brings richer visual appearances while preserving key semantics.
    
    \item We collect a fractal image dataset 
    and blend into the hybrid images. This improves the overall structural complexity of the augmented images and helps to avoid over-fitting on generated images, thus resulting in better generalization.


    
    \item Extensive experiments on seven datasets for various tasks including \emph{general classification, fine-grained classification, adversarial robustness, transfer learning, and data scarcity} demonstrate the superior performance of our proposed method compared to the existing SOTA image augmentation techniques.

\end{itemize}


\begin{figure*}[t]
    \centering
    \includegraphics[width=1\textwidth]{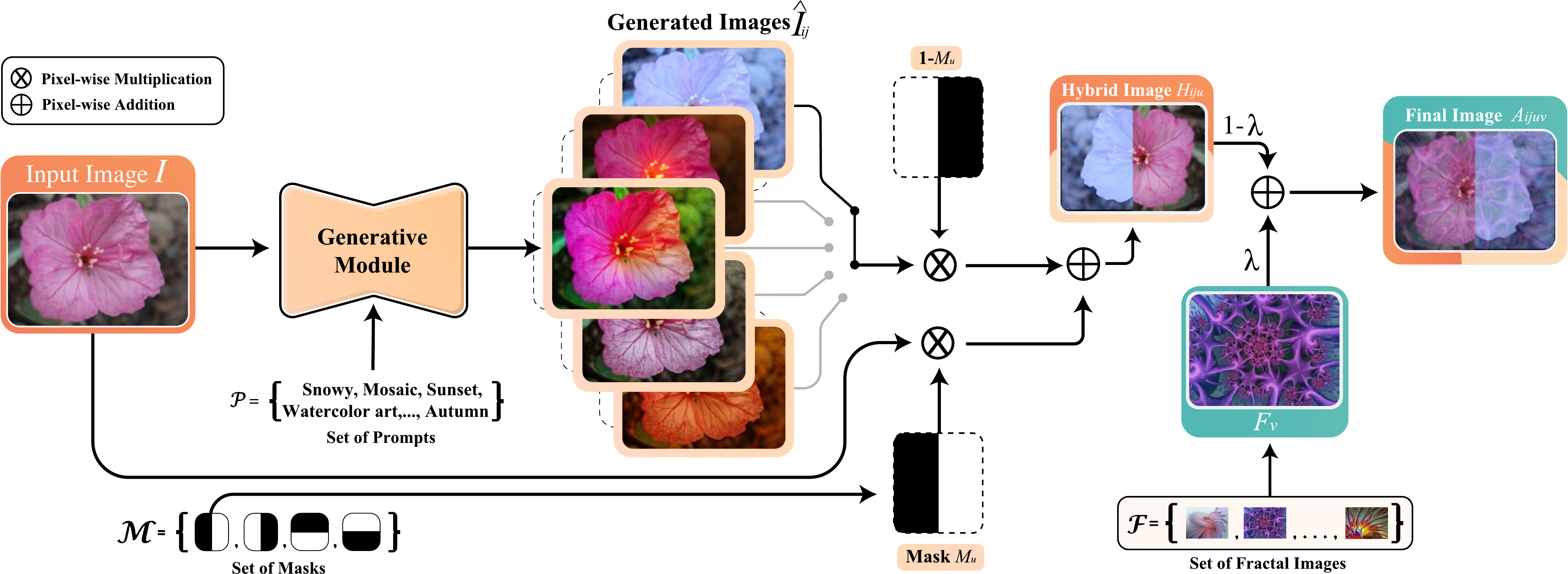}
    \vspace{-1.0em}
    \caption{ \textbf{Architecture of the proposed \textsc{DiffuseMix} approach}. An input image and a randomly selected prompt are input to a diffusion model to obtain a generated image. Input and generated images are concatenated using a binary mask to obtain a hybrid image.  A random fractal image is finally blended with this hybrid image to obtain the augmented image.}
    \label{fig:diffuseMixPipeline}
    \vspace{-1.5em}
\end{figure*}



\section{Related Work}
\label{sec:relatedWork}
Data augmentation has become indispensable in enhancing the diversity of training datasets, thereby mitigating the risks of overfitting. Traditional approaches employed strategies—such as horizontal and vertical translations, affine transformations, scaling, and squeezing—when training a model. This not only improves the performance but also improves the generalization of the model on test datasets. 

\vspace{2pt}
\noindent \textbf{Diffusion Models for Augmentation:} Recently, several researchers have explored the possibility of data augmentation with diffusion models. \citet{azizi2023synthetic}  proposed the utilization of fine-tuned text-to-image diffusion models on ImageNet classification, revealing that augmenting the training set with these synthetic samples may boost classification performance. Similarly, \citet{trabucco2023effective} investigated diffusion models to create more diverse and semantically varied datasets, aiming to improve outcomes in tasks such as image classification. \citet{li2023synthetic} further explored diffusion models-based augmentation for knowledge distillation without real images. 
\vspace{2pt}

\noindent \textbf{Image Mixing Augmentation:} Image mixing is a prominent class of augmentation methods for training robust CNN models \cite{liang2023miamix, yan2023locmix, mensink2023infinite, chen2023rankmix}. Some of these methods include Mixup, CutMix, and AugMix. Mixup \cite{zhangmixup} generates synthetic images by linearly interpolating pixel values from two randomly selected images. In contrast, CutMix \cite{yun2019cutmix} involves pasting a random patch from one image onto another. AugMix \cite{hendrycksaugmix} employs a stochastic combination of data augmentation operations on an input image. SaliencyMix \cite{uddin2020saliencymix} utilizes saliency maps to concentrate the augmentation on the image's most vital regions, ensuring overall image integrity. Manifold Mixup \cite{verma2019manifold} enhances representation by interpolating network hidden states during training. This entails blending two hidden states with a random weight to produce an interpolated manifold-based hidden state. PuzzleMix \cite{kim2020puzzle}, an improvement over the traditional mixup, factors in image saliency, and local statistics during image blending. This method segments an image into patches, allocates weights based on saliency and local statistics, and merges patches from different images in accordance with their weights. PixMix \cite{hendrycks2022pixmix} have studied mixing of input images with fractal and feature visualization images to improve ML safety measures. A detailed summary of several image-mixing-based methods along with their components and application tasks is provided in Table \ref{table:compar_study}.
\vspace{2pt}

\noindent \textbf{Automated Augmentation:} AutoAugment \cite{cubuk2018autoaugment}, for instance, employed reinforcement learning to pinpoint optimal data augmentation policies, while RandAugment \cite{cubuk2020randaugment} integrates a suite of random data augmentation operations to improve model generalization. AdaAug\cite{cheung2021adaaug} is proposed to efficiently learn adaptive augmentation policies in a class-dependent and potentially instance-dependent manner. 
\par
In contrast to previous methods, our approach emphasizes the concatenation of original and generated images, using a pre-defined library of conditional prompts. The obtained hybrid images are blended with fractal images to further improve the overall performance. 

\newcommand{\our}{\textsc{DiffuseMix}}

\section{\textsc{DiffuseMix}}
\label{sec:method}

\subsection{Background and Overview}

Existing image-mixing-based methods may induce label ambiguity by placing one image on top of the other and consequently overlapping either some portions of the object or its context  \cite{cutmix, kang2023guidedmixup}. In contrast, the core idea of \textsc{DiffuseMix} is to concatenate a portion of the original image with its counterpart generated image in such a way that basic image semantics are preserved while providing diverse object details and contexts for better augmentation.
\par
\begin{figure}[t]
\includegraphics[width=0.48\textwidth]{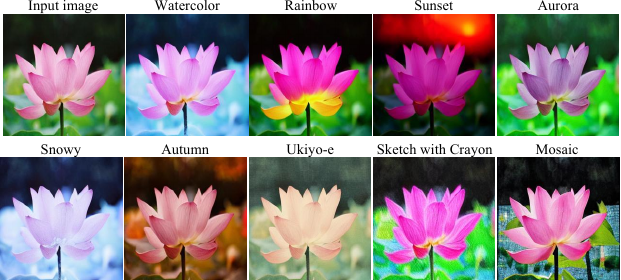}
\vspace{-1.9em}
\caption{A set of \emph{bespoke conditional prompts}  are used to obtain generated images preserving important features and adding rich visual appearance to the input images.}
\label{fig:prompts_variation}
\vspace{-1.3em}
\end{figure}
The proposed method as illustrated in Figure \ref{fig:diffuseMixPipeline}  comprises of three pivotal steps: 
\textbf{\emph{generation}, \emph{concatenation}}, and \textbf{ \emph{fractal blending}}. Firstly, conditional prompts are used with a diffusion model to obtain a generative counterpart of the input image. Then, a portion of the original image is concatenated with the rest of the portion taken from the generated image forming a hybrid image. This step is to ensure that the training network always has access to the original data along with the generated one. Subsequently, a random fractal image is blended into the hybrid image to obtain the final training image with a diverse structure. 
Blending fractal images has proven to be effective towards ML safety ~\cite{hendrycks2022pixmix, hendrycks2021unsolved}. In our work, we study the effectiveness of blending fractal images mainly towards improved performance.




\subsection{Method}
The proposed \textsc{DiffuseMix} is an effective data augmentation technique which can be used to enhance the robustness and generalization of the deep learning models. 
Formally,  $I_i \in \mathbb{R}^{h\times w \times c}$ is an image from the training dataset, $\mathcal{D}$\textsubscript{mix}$(\cdot): \mathbb{R}^{h\times w \times c} \to \mathbb{R}^{h\times w \times c}$ denotes our data augmentation method. 
To obtain the final augmented image $A_{ijuv}$, input image $I_i$ goes through proposed generation using prompt $p_j$, concatenation using mask $M_u$, and blending using fractal image $F_v$. The overall augmentation process, as also seen in Algorithm \ref{algo:diffuseMix}, can be represented as $ A_{ijuv} = \mathcal{D}$\textsubscript{mix}$(I_i,p_j,M_u,F_v,\lambda)$.



\vspace{5pt}
\noindent \textbf{Generation:}
Our generation step $\mathcal{G}(.)$ consists of a \textit{pre-trained} diffusion model that takes a prompt $p_j$ from a predefined set of $k$ prompts, $P = \{p_1, p_2, \dots, p_k\}$ where $j \in [1, k]$ along with the input image $I_i$ and produces an augmented counterpart image $\hat{I}_{ij}$. Image editing process in conventional diffusion models is often open-ended and guided by text prompts to obtain diverse image-to-image or text-to-image translations.
In our case, as the goal is to achieve a slightly modified but not too different version of $I_i$, \emph{filter-like} prompts are curated in $\mathcal{P}$ which do not alter the image drastically. Examples of the prompts used in \textsc{ DiffuseMix} are shown in Figure \ref{fig:prompts_variation}.
The overall generation step can be represented as: 
   $ \hat{I}_{ij} = \mathcal{G}(I_i,p_j)$, where $p_j$ is a randomly selected prompt.


\vspace{5pt}




\noindent \textbf{Concatenation:}
We concatenate a portion of the original input image $I_i$ with its counterpart generated image $\hat{I}_{ij}$  using a randomly selected mask $M_u$ from the set of masks  to create a hybrid image $H_{iju}$: 
\begin{equation}
    H_{iju}= (\hat{I}_{ij} \odot M_u) + (I_{i} \odot (\bold{1}-M_u)).
    \label{equ:one}
\end{equation}
The mask $M_u$ consists of zeros and ones only and $\odot$ is a pixel-wise multiplication operator.
The set of masks contains four kinds of masks including horizontal, vertical and flipped versions. 
Such masking ensures the availability of the semantics of the input image to the learning network while reaping the benefits of the generated images.



\vspace{5pt}
\noindent \textbf{Fractal Blending:}
A fractal image dataset\footnote{Examples of fractal images are provided in \textcolor{customviolet}{Appendix 8.}}  $\mathcal{F}$ is collected and used for inducing structural variations in the hybrid images.
A randomly  selected fractal image $F_v \in \mathcal{F}$ is blended to the hybrid image $H_{iju}$ with a blending factor $\lambda$ as:
\begin{equation}
    A_{ijuv}= \lambda F_v + (1-\lambda) H_{iju},
     \label{equ:two}
\end{equation}

\noindent where $\lambda$ is the blending factor.
This results in the final augmented image $A_{ijuv}$    used to train or fine-tune a deep learning model. 
The overall augmentation process of \textsc{DiffuseMix} can be represented as:
\begin{equation}
    A_{ijuv} =(1 - \lambda)(I_i \odot M_u + \hat{I}_{ij} \odot (\bold{1}-M_u))+\lambda F_v,
    \label{eq:masking}
\end{equation}

\begin{algorithm}[t]
\caption{ \textsc{DiffuseMix}}
\begin{algorithmic}[1]
\Require $I_i \in \mathcal{D}$ training images dataset, $m$: number of augmented images,
 $p_j \in \mathcal{P}$ set of prompts, $M_u \in \mathcal{M}$ set of masks, $F_v \in \mathcal{F}$ library of fractal images, $\lambda$: blend ratio 
\Ensure $\mathcal{D}'$: $m$ Augmented images  
\State $\mathcal{D}' \gets \emptyset$ 
\For{each image $I_i$ in $\mathcal{D}$}
\For{$a$ in  $\{1:m\}$}
    \State Randomly select prompt $p_j$ from $\mathcal{P}$ 
    \State Generate image: $\hat{I}_{ij} \gets \mathcal{G}(I_i, p_j)$ 
       \State Randomly select mask $M_u$ from $\mathcal{M}$
            \State Hybrid image: $H_{iju} \gets$  $M_u \odot I_i+(1-M_u) \odot \hat{I}_{ij}$ 
            \State  Randomly select $F_v$ from $\mathcal{F}$
            \State Blended image: $A_{ijuv} \gets (1-\lambda) H_{iju}+ \lambda F_v$ 
            \State Add $A_{ijuv}$ to $\mathcal{D}'$
    \EndFor
\EndFor

\State \textbf{return} $D'$
\end{algorithmic}

\label{algo:diffuseMix}
\end{algorithm}

\section{Experiments and Results}
\label{sec:experiment}
In this section, we present the experimental details, datasets used to evaluate our approach, and analyses of the results.


\vspace{2pt}
\noindent{\textbf{Datasets.}} 
To provide comparisons with existing studies on image augmentation  \cite{cutmix, mixup, kang2023guidedmixup, uddin2020saliencymix, hendrycksaugmix, cheung2021adaaug, cubuk2020randaugment, autoaugment, kim2020puzzle, kim2020co, huang2021snapmix, lim2019fast, verma2019manifold}, we evaluate our approach on several \emph{general image classification} and \emph{fine-grained image classification} datasets.
In the general image classification category, we employ three datasets including ImageNet \cite{imagenet}, CIFAR100 \cite{krizhevsky2009learning} and Tiny-ImageNet-200 \cite{le2015tiny}.
In fine-grained image classification category, we employ four datasets including Oxford-$102$ Flower \cite{nilsback2008automated}, Stanford Cars \cite{krause20133d}, Aircraft~\cite{maji2013fine}, and Caltech-UCSD Birds-200-2011 (CUB)  \cite{wah2011caltech}.
These datasets offer a diverse array of scenarios where images contain a wide range of objects such as plants and animals in various scenes, textures, transportation modes, human actions, satellite imagery, and general objects.


\begin{table*}[t]
\renewcommand{\arraystretch}{1.1}
    \setlength{\tabcolsep}{3.3pt}
    \begin{minipage}{0.58\textwidth}
        \caption{Top-1 and Top-5 accuracy on \textbf{\emph{general classification task}} of PreactResNet-18 trained from scratch for 300 epochs following the results of \citet{kang2023guidedmixup}. Extended table can be seen in \textcolor{customviolet}{Appendix 7 Table 12.}}
        \vspace{-7pt}
    \scalebox{0.89}{
    
\begin{tabular}{p{3.8cm}p{1.6cm}<{\centering}p{1.6cm}<{\centering}|p{1.6cm}<{\centering}p{1.6cm}<{\centering}}
\hline\thickhline
\rowcolor{mygray}
& \multicolumn{2}{c|}{Tiny-ImageNet-200} & \multicolumn{2}{c}{CIFAR-100} \\
\cline{2-5} 
\rowcolor{mygray}
\multirow{-2}{*}{Method} & Top-1 (\%) & Top-5 (\%) & Top-1 (\%) & Top-5 (\%) \\
\hline
Vanilla\textsubscript{\textcolor{blue}{(CVPR'16)}} \cite{resnet} & 57.23 & 73.65 & 76.33 & 91.02 \\
SaliencyMix\textsubscript{\textcolor{blue}{(ICLR'21)}} \cite{uddin2020saliencymix} & 56.54 & 76.14 & 79.75 & 94.71 \\
Guided-SR\textsubscript{\textcolor{blue}{(AAAI'23)}} \cite{kang2023guidedmixup} & 55.97 & 74.68 & 80.60 & 94.00 \\
PuzzleMix\textsubscript{\textcolor{blue}{(ICML'20)}} \cite{kim2020puzzle}  & 63.48 & 75.52 & 80.38 & 94.15 \\
Co-Mixup\textsubscript{\textcolor{blue}{(ICLR'21)}} \cite{kim2020co} & 64.15 & - & 80.15 & - \\
Guided-AP\textsubscript{\textcolor{blue}{(AAAI'23)}} \cite{kang2023guidedmixup} & 64.63  & 82.49  & 81.20 & 94.88   \\ \hline
\textbf{\textsc{DiffuseMix}}  & \textbf{65.77} & \textbf{83.66} & \textbf{82.50}  & \textbf{95.41} \\
\bottomrule
\end{tabular}

    }
    \label{tab:tiny_image_net_cifar100}
    \end{minipage}
    \hfill
    \begin{minipage}{0.403\textwidth} 
        \caption{Top-1 / Top-5 performance on ImageNet-1K dataset benchmark when trained on ResNet-50 for 100 epochs for 
\textbf{\emph{general classification task}}. An extended version of this table is provided in \textcolor{customviolet}{Appendix 7 Table 13.}}
\vspace{-7pt}
        \setlength{\tabcolsep}{3.5pt}
    \resizebox{0.963\columnwidth}{!}{
    \begin{tabular}{lcc}
    \hline\thickhline
    \rowcolor{mygray} 
Method & Top-1 (\%) & Top-5 (\%) \\

\hline
Vanilla\textsubscript{\textcolor{blue}{(CVPR'16)}} \cite{resnet} & 75.97 & 92.66     \\
PixMix\textsubscript{\textcolor{blue}{(CVPR'22)}} \cite{hendrycks2022pixmix} & 77.40 & - \\
PuzzleMix\textsubscript{\textcolor{blue}{(ICML'20)}} \cite{kim2020puzzle} & 77.51 & 93.76 \\
GuidedMixup\textsubscript{\textcolor{blue}{(AAAI'23)}} \cite{kang2023guidedmixup} & 77.53 & 93.86 \\
Co-Mixup\textsubscript{ \textcolor{blue}{(ICLR'21)}} \cite{kim2020co} & 77.63 & 93.84 \\ 
YOCO\textsubscript{\textcolor{blue}{(ICML'22)}} \cite{han2022yoco} & 77.88 & - \\
\hline
\textbf{\textsc{DiffuseMix}}  & \textbf{78.64} & \textbf{95.32} \\
\bottomrule
    \end{tabular}
    }

    \label{tab:image_net}
    \end{minipage}
\end{table*}

\vspace{2pt}
\noindent{\textbf{Implementation Details.}} We utilize InstructPix2Pix \cite{brooks2023instructpix2pix} diffusion model to generate images with the help of our introduced textual library. For the generation of Mask $M$ in Eq. 1, a template image is divided into two equal parts, either horizontally or vertically. Randomly, one half is turned on and the second half is turned off. In all experiments \footnote{Following AugMix \cite{hendrycksaugmix}, we employ JS-divergence during training.}, $\lambda=0.20$ is used for blending the fractal image in Eq. \eqref{equ:two} \& \eqref{eq:masking}. Analysis on $\lambda$ values is provided in  \textcolor{customviolet}{Appendix 1.}






\vspace{5pt}
\noindent{\textbf{Textual Prompt Selection.}} In order to ensure that only appropriate prompts are applied, a bespoke textual library of \textit{filter-like} global visual effects is
predefined: 
\textcolor{customviolet}{\emph{`autumn', `snowy', `sunset', `watercolor art', `rainbow', `aurora', `mosaic',`ukiyo-e', and `a sketch with crayon'}}.  These prompts are selected because of their generic nature and applicability to a wide variety of images. Secondly, these do not alter the image structure significantly while producing a global visual effect in the image.  Each prompt in the textual library is appended with a template `A transformed version of image into $prompt$' to form a particular input to the diffusion model. Examples of images generated through these prompts are shown in Figure \ref{fig:prompts_variation}. Additional visual examples and discussions on appropriate prompt selection are provided in \textcolor{customviolet}{Appendix 3.}



\begin{figure}[t]
    \includegraphics[width=0.46\textwidth]{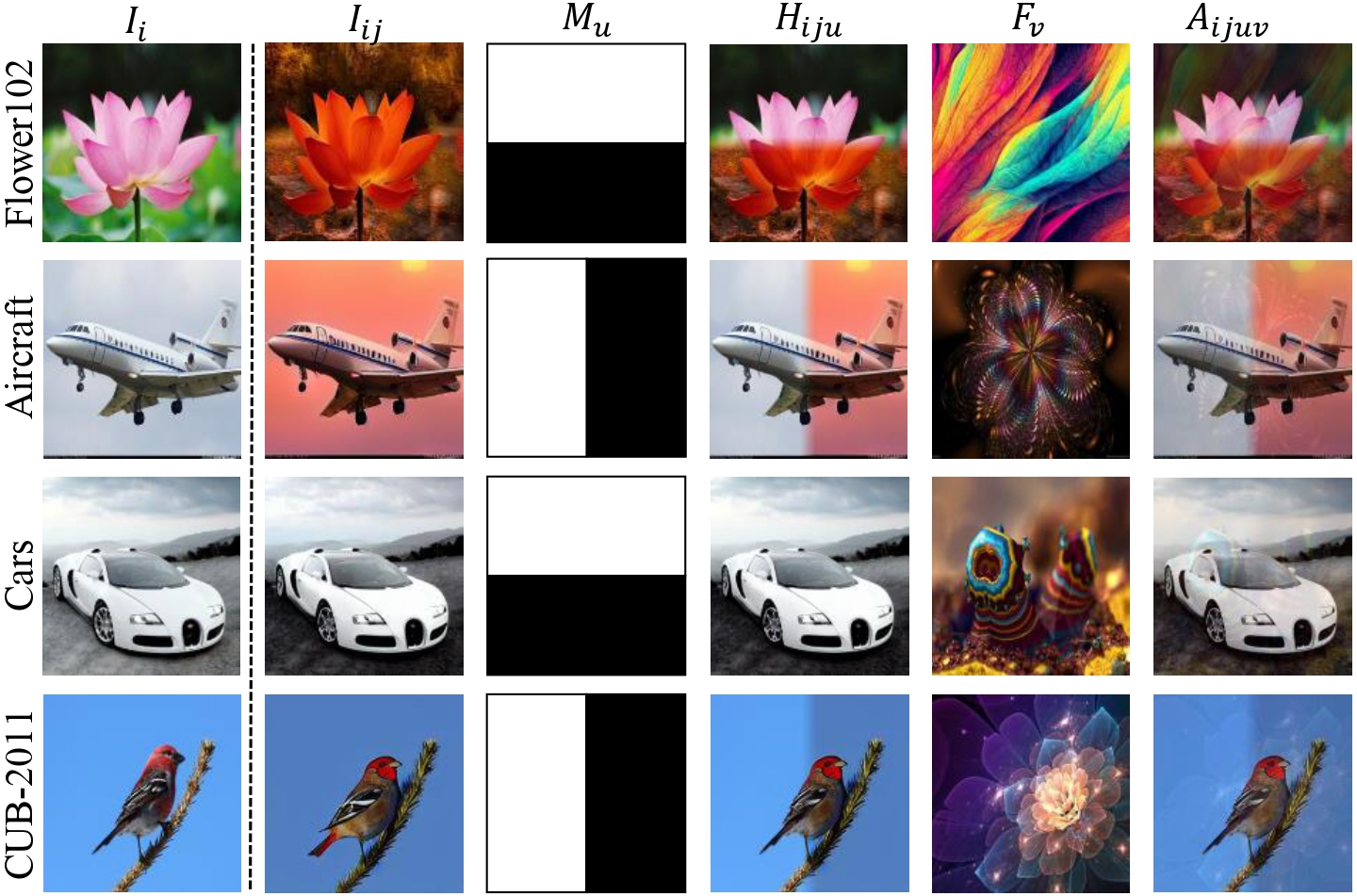} %
    \vspace{-5pt}
    \caption{Example images from different stages of \textsc{DiffuseMix}: input image ($I_i$), generated image ($\hat{I}_{ij}$), mask ($M_u$), hybrid image ($H_{iju}$), fractal image ($F_v$), and final augmented image ($A_{ijuv}$).}
    \label{fig:stages}
    \vspace{-1.0em}
\end{figure}

\vspace{5pt}
\noindent \textbf{Visualizing Intermediate Steps.} Figure \ref{fig:stages} depicts images obtained in each step of \textsc{DiffuseMix}.
It can be observed that \textsc{DiffuseMix} yields a broader spectrum of augmented images derived from the training set. These images contain full object with no portions omitted and provide suitable variations for training.


\section{Performance Evaluation}

\subsection{General Classification }
General Classification (GC) can be a notable way to quantify the effectiveness of an image augmentation method. A higher GC accuracy would mean that an augmentation method successfully provided plausible data variations to improve the learning process. 
We evaluate our approach for the GC task on three challenging datasets including Tiny-ImageNet-200, CIFAR-100, and ImageNet.
Similar to the existing SOTA methods, PreActResNet18 network is employed for the experiments on Tiny-ImageNet-200 and CIFAR-100 datasets \cite{huang2021snapmix, kang2023guidedmixup, kim2020co, uddin2020saliencymix}, while ResNet$50$ is employed for ImageNet dataset \cite{kim2020puzzle, hendrycksaugmix, cutmix}.


In Table \ref{tab:tiny_image_net_cifar100} Top-1 and Top-5 accuracy on Tiny-ImageNet and CIFAR-$100$ datasets is compared with the existing SOTA methods.
The proposed \textsc{DiffuseMix} technique demonstrates better performance gains compared to the existing image augmentation approaches.
On Tiny-ImageNet dataset,
compared to Vanilla model, our \textsc{DiffuseMix} results in notable Top-1 and Top-5 accuracy gains of \numbersBlue{$8.54\%$} and \numbersBlue{$10.01\%$}.
Moreover, compared to the second best performer, Guided-AP \cite{kang2023guidedmixup},  Top-1 and Top-5 accuracy gains of our approach are \numbersBlue{$1.14\%$} and \numbersBlue{$1.17\%$}.  
Similar trends are observed on CIFAR$100$ dataset, where \textsc{DiffuseMix} demonstrates Top-1 and Top-5 accuracy gains of  \numbersBlue{$6.17\%$} and \numbersBlue{$4.39\%$} over vanilla and \numbersBlue{$1.3\%$} and \numbersBlue{$0.53\%$} over Guided-AP.

We also evaluate \textsc{DiffuseMix} on large-scale ImageNet dataset offering more challenging scenarios where existing SOTA methods \cite{cutmix,hendrycksaugmix,kim2020puzzle} only report Top-1 accuracy. Compared to the second best performer PuzzleMix \cite{kim2020puzzle}, our approach demonstrates a performance gain of \numbersBlue{$1.13\%$} (Table \ref{tab:image_net}). Compared to Vanilla implementation,  our approach demonstrates a performance gain of \numbersBlue{$2.95\%$}.
The GC results on these challenging and diverse benchmark datasets highlight the effectiveness of \textsc{DiffuseMix} in enabling better learning. It also suggests its capability to combat overfitting and achieve better generalization.







\subsection{Adversarial Robustness}
Following existing SOTA methods \cite{kim2020puzzle,cutmix,hendrycksaugmix, verma2019manifold,uddin2020saliencymix,hong2021stylemix,kim2020co,kang2023guidedmixup}, we evaluate the robustness of our approach against adversarial attacks and input perturbations.
In these experiments, fast adversarial training \cite{fast_imagenet} is adapted to create adversarially perturbed input images. The goal of these experiments is to evaluate whether an augmentation approach can demonstrate better resilience against adversarial attacks. FGSM \cite{fast_imagenet} error rates are computed to evaluate the performance against adversarial attacks. 

As shown in Table \ref{tab:fgsm}, \textsc{DiffuseMix} demonstrates an error rate of \numbersBlue{$17.38\%$} on CIFAR-100, which is lower than all compared methods. PuzzleMix is the second best performer obtaining \numbersBlue{$19.62\%$} error rate. Similarly on Tiny ImageNet-200, \textsc{DiffuseMix} outperforms SOTA by a notable margin obtaining \numbersBlue{$34.53\%$} error rate while PuzzleMix remained the second best performer with \numbersBlue{$36.52\%$} error rate. These results demonstrate that even under adversarial perturbations, \textsc{DiffuseMix} remains resilient surpassing the performance of existing SOTA approaches.

\begin{table}
\renewcommand{\arraystretch}{1.1}
\caption{FGSM error rates on CIFAR-100 and Tiny-ImageNet-200 datasets for PreactResNet-18, following \cite{kim2020puzzle}.}
\vspace{-7pt}
\setlength{\tabcolsep}{7pt}
\centering
\resizebox{0.97\columnwidth}{!}{
\begin{tabular}{lcc}
\hline\thickhline
\rowcolor{mygray}
& \multicolumn{2}{c}{FGSM Error Rates (\%)} \\
\rowcolor{mygray}
\multirow{-2}{*}{Method} & CIFAR-100 & Tiny-ImageNet-200 \\ 
\midrule 
Vanilla\textsubscript{\textcolor{blue}{(CVPR'16)}} \cite{resnet} & 23.67 & 42.77 \\
Mixup \textsubscript{ \textcolor{blue}{(ICLR'18)}} \cite{mixup} & 23.16 & 43.41 \\
Manifold\textsubscript{ \textcolor{blue}{(ICML'19)}} \cite{verma2019manifold} & 20.98 & 41.99 \\
CutMix \textsubscript{ \textcolor{blue}{(ICCV'19)}} \cite{cutmix} & 23.20 & 43.33 \\
AugMix \textsubscript{ \textcolor{blue}{(ICLR'20)}} \cite{hendrycksaugmix} & 43.33 & - \\
PuzzleMix\textsubscript{\textcolor{blue}{(ICML'20)}} \cite{kim2020puzzle}  & 19.62 & 36.52 \\
\hline
\textbf{\textsc{DiffuseMix}}  & \textbf{17.38} & \textbf{34.53} \\
\bottomrule
\end{tabular}
}
\vspace{-5pt}
\label{tab:fgsm}
\end{table}




\par
\subsection{Fine-Grained Visual Classification }
Compared to the general classification, the task of Fine-Grained Visual Classification (FGVC) is notably challenging since it is difficult for a learning model to identify subtle differences between two different objects belonging to the same general class. 
A robust image augmentation technique should preserve these subtle albeit critical details for the learning model to successfully train on the fine-grained classification task. To evaluate \textsc{DiffuseMix} on this task, we conduct experiments on three datasets including CUB \cite{nilsback2008automated}, Stanford Cars \cite{krause20133d}, and Aircraft~\cite{maji2013fine} utilizing ResNet-50 \cite{resnet} network.


As shown in Table \ref{tab:fgvc_acc},  \textsc{DiffuseMix} significantly enhances the generalization capability of ResNet-50, yielding \emph{superior} performances on all benchmark datasets.
Specifically, \textsc{DiffuseMix} outperforms widely-acknowledged methods such as Mixup and CutMix, with notable margins. On CUB dataset, \textsc{DiffuseMix} achieved an accuracy of \numbersBlue{$79.37\%$}, which is a significant leap from the \numbersBlue{$65.50\%$}  Vanilla accuracy. It also outperformed recent methods including  Guided-AP \numbersBlue{$77.08\%$},  and PuzzleMix \numbersBlue{$72.83\%$}. Similarly, on the Aircraft dataset, \textsc{DiffuseMix} obtained an accuracy of \numbersBlue{$85.76\%$} surpassing the second best, GuidedMixup \numbersBlue{$84.32 \%$}. 
The Stanford Cars dataset further validates the remarkable performance of \textsc{DiffuseMix} registering \numbersBlue{$91.26\%$} accuracy. On the same dataset, other methods such as SnapMix and PuzzleMix achieved \numbersBlue{$90.10\%$} and \numbersBlue{$89.68\%$} respectively. The consistent performance gains on various challenging datasets iterate the effectiveness of \textsc{DiffuseMix} in retaining critical salient information necessary to perform fine-grained classification.



\begin{table}
\renewcommand{\arraystretch}{1.1}
\caption{Top-1 (\%) performance comparison on \textbf{\emph{fine-grained task}} of ResNet-50. 
Extended comparisons are provided in \textcolor{customviolet}{Appendix 7 Table 14.}}
\label{tab:fgvc_acc}
\vspace{-7pt}
\centering
\resizebox{0.963\columnwidth}{!}{
\begin{tabular}{p{3.6cm}p{1.1cm}<{\centering}p{1.2cm}<{\centering}p{1.1cm}<{\centering}}
\hline\thickhline
\rowcolor{mygray} 

Method & Birds & Aircraft & Cars 
\\ \hline

Vanilla\textsubscript{\textcolor{blue}{(CVPR'16)}}  ~\cite{resnet} & 65.50 & 80.29 & 85.52 \\ 

RA\textsubscript{\textcolor{blue}{(NIPS'20)}}  \cite{cubuk2020randaugment} & - & 82.30 & 87.79 \\
AdaAug\textsubscript{\textcolor{blue}{(ICLR'22)}}  \cite{cheung2021adaaug} & - & 82.50 & 88.49 \\

Mixup\textsubscript{\textcolor{blue}{(ICLR'18)}} ~\cite{mixup} & 71.33 & 82.38 & 88.14 \\
CutMix\textsubscript{\textcolor{blue}{(ICCV'19)}}~\cite{cutmix} & 72.58 & 82.45 & 89.22 \\
SnapMix\textsubscript{\textcolor{blue}{(AAAI'21)}} ~\cite{huang2021snapmix} & 75.53 & 82.96 & 90.10 \\
PuzzleMix\textsubscript{\textcolor{blue}{(ICML'20)}}  ~\cite{kim2020puzzle} & 74.85 & 82.66 & 89.68 \\
Co-Mixup\textsubscript{\textcolor{blue}{(ICLR'21)}}  ~\cite{kim2020co} & 72.83 & 83.57 & 89.53 \\
Guided-AP\textsubscript{\textcolor{blue}{(AAAI'23)}}  ~\cite{kang2023guidedmixup} & 77.08  & 
84.32  & 90.27  \\
\hline
\textbf{\textsc{DiffuseMix}}  & \textbf{79.37} & \textbf{85.76} & \textbf{91.26} \\
\bottomrule
\end{tabular}
}

\end{table}
\subsection{Transfer Learning}
Transfer learning or fine-tuning is a widely used way of customizing large architectures with limited computational resources as well as for quick experiments. Most image-mixing-based augmentation methods  \cite{cutmix,mixup,kang2023guidedmixup,uddin2020saliencymix,hendrycksaugmix,kim2020puzzle,kim2020co,huang2021snapmix,kang2023guidedmixup,hendrycksaugmix,verma2019manifold} have not reported performance in this important scenario.  Nevertheless, we evaluate the performance of our approach on fine-tuning the baseline model using three different datasets including Flower102, Aircraft, and Stanford Cars on  ImageNet-pretrained ResNet-50 provided by PyTorch and report results in Table  
\ref{tab:fine_tunning}.

For the Flower102 dataset, \textsc{DiffuseMix} achieved an accuracy of \numbersBlue{ $98.02\%$}, which is higher than the second best method, AdaAug \numbersBlue{$97.19\%$}. A similar trend is observable in the Aircraft \numbersBlue{$85.65\%$} and Cars \numbersBlue{ $93.17\%$} datasets. 
We also observe that the accuracy obtained by \textsc{DiffuseMix} with fine-tuning (Table \ref{tab:fine_tunning}) is comparable to the performance when training is done from scratch (Table \ref{tab:fgvc_acc}). Since fine-tuning consumes significantly less computational resources compared to training from scratch, this experiment elaborates the practical significance of \textsc{DiffuseMix}.




\begin{SCfigure*}
    \centering
    \begin{subfigure}[b]{0.54\columnwidth} 
        \centering
        \includegraphics[width=\textwidth,  keepaspectratio]{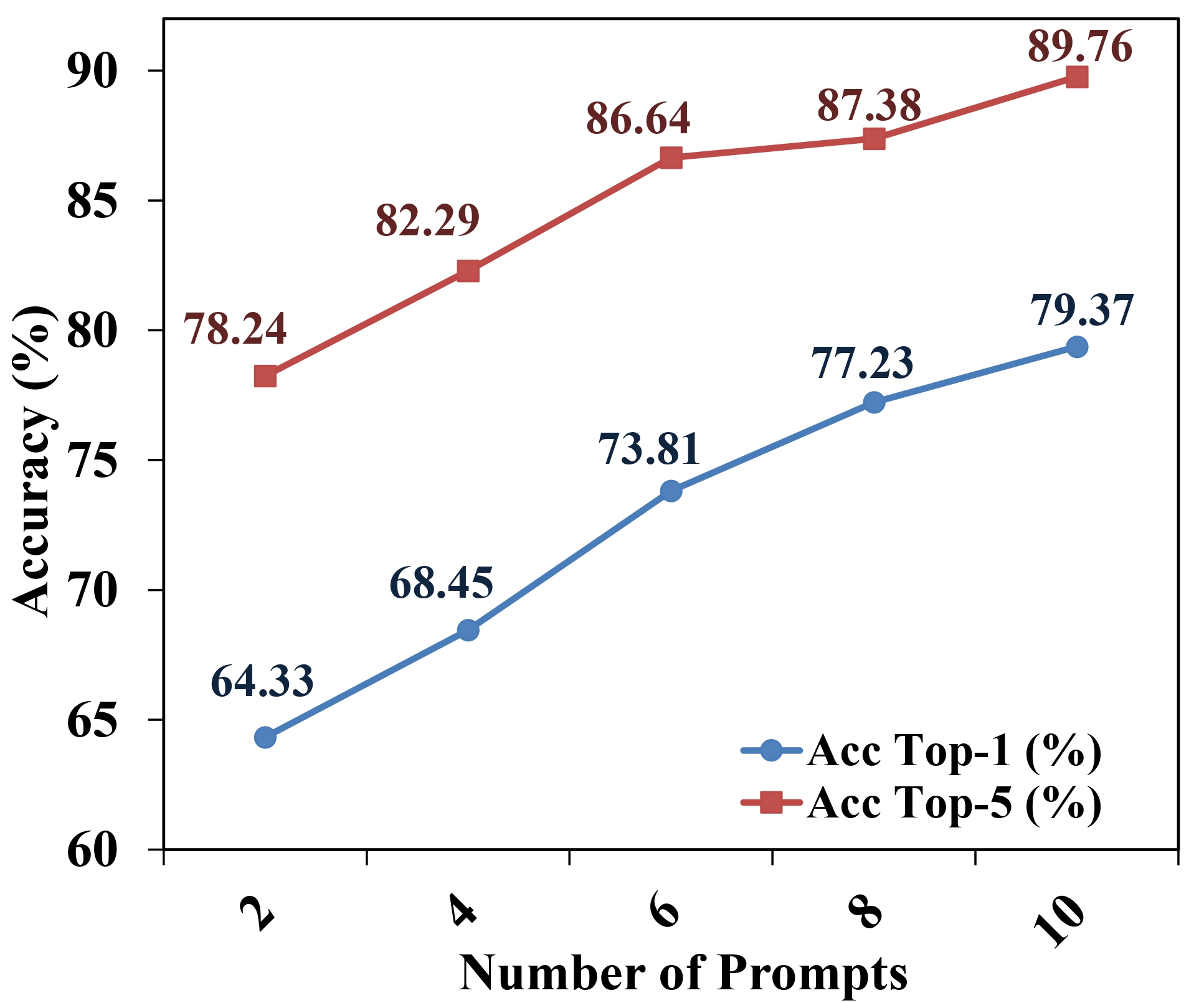}
        \caption{CUB-200-2011}
    \end{subfigure}%
     \begin{subfigure}[b]{0.54\columnwidth} 
        \centering
        \includegraphics[width=\textwidth,  keepaspectratio]{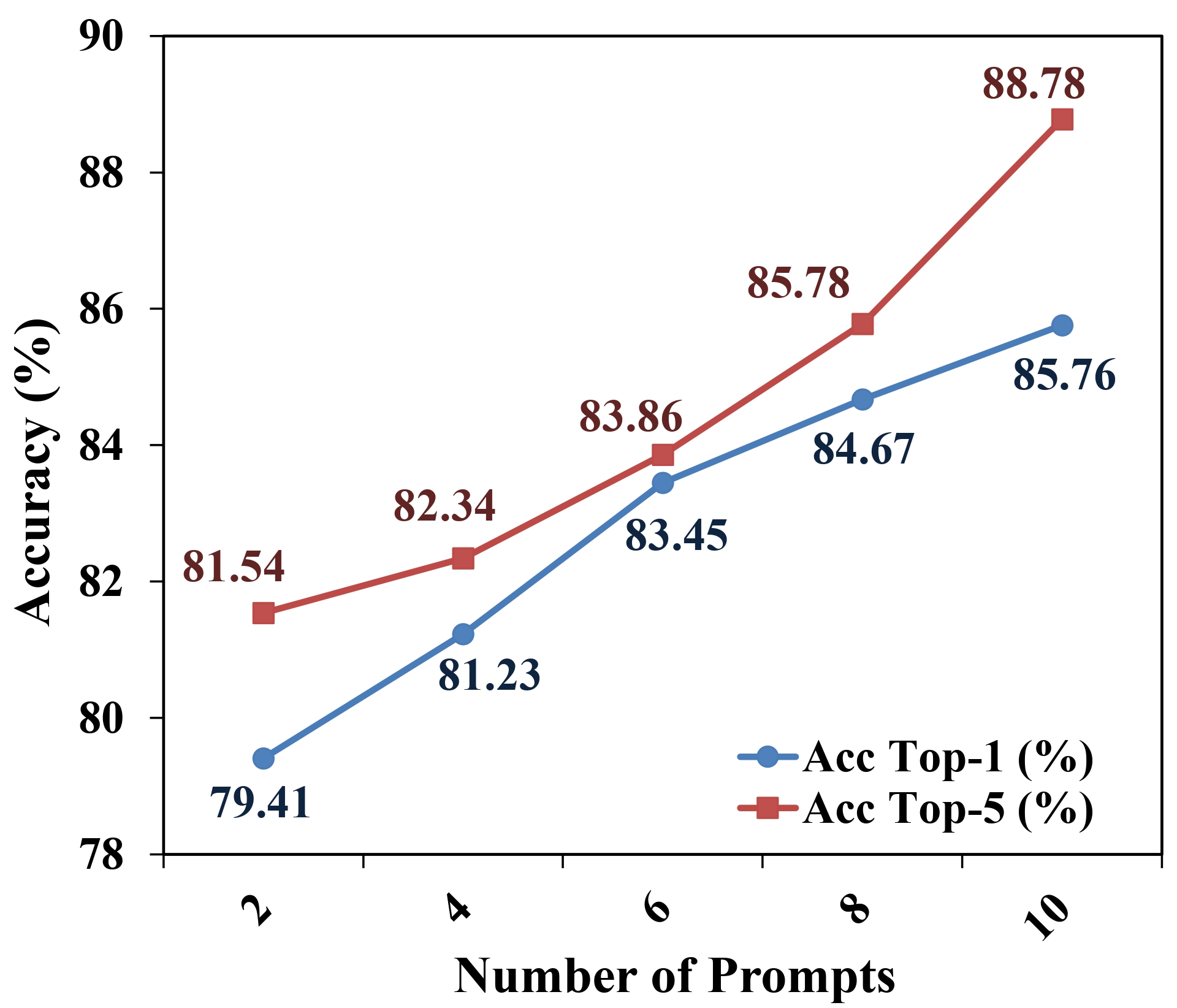}
        \caption{FGVC Aircraft}
    \end{subfigure}%
    \begin{subfigure}[b]{0.54\columnwidth}
        \centering
        \includegraphics[width=\textwidth,  keepaspectratio]{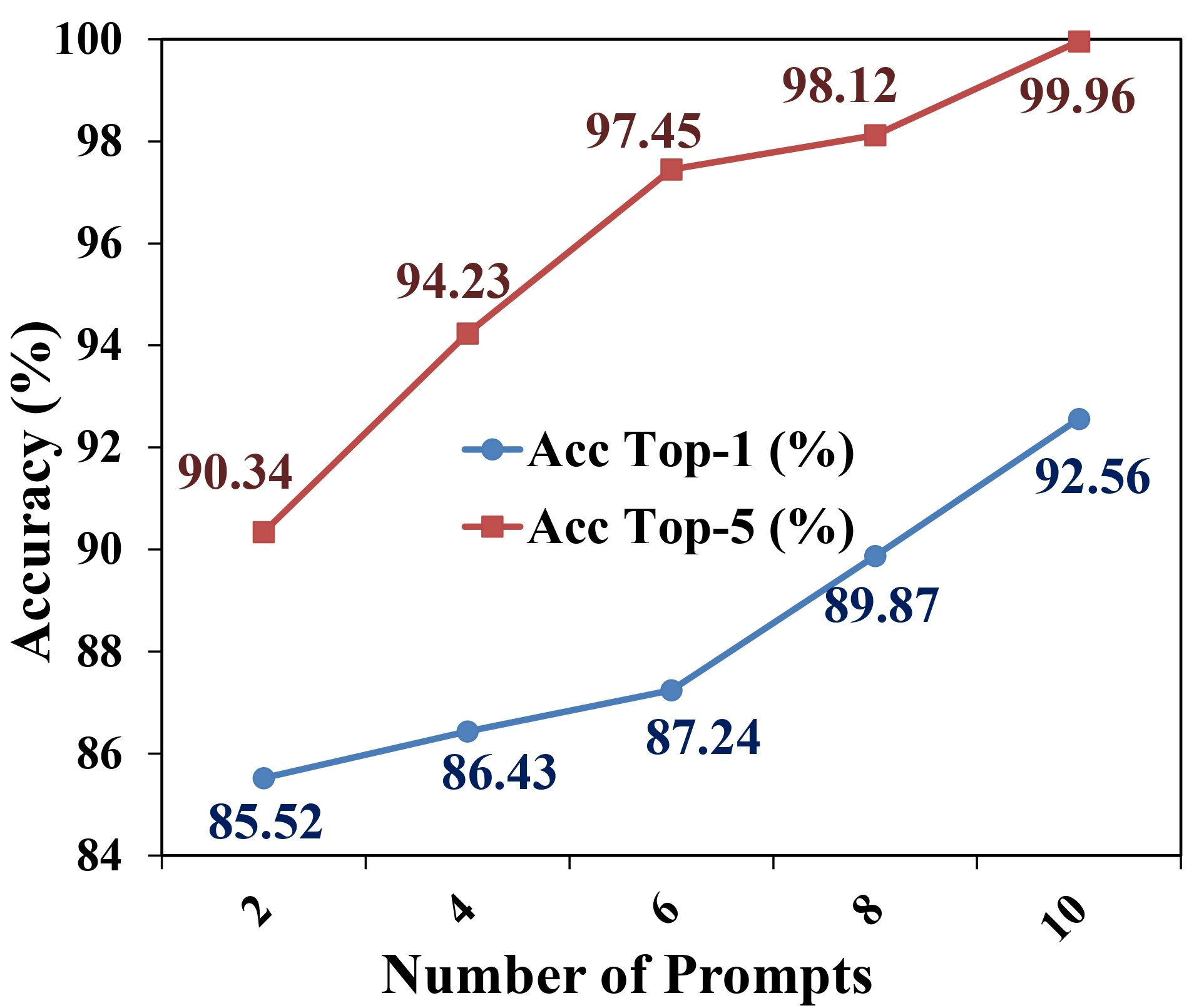}
        \caption{Stanford Cars}
    \end{subfigure}
    \caption{Effect of the number of prompts on overall performance.
    A detailed ablation study showcases the gains in Top-1 (\%) and Top-5 (\%) accuracy across CUB Birds-200, Aircraft, and Stanford Cars datasets with an increase in the number of prompts in \textsc{DiffuseMix}.}
    \label{fig:numofprompts}
\end{SCfigure*}

\subsection{Data Scarcity} Data scarcity is one of the major issues when training deep neural networks. 
If the training examples per class are limited, deep networks may not learn meaningful patterns by leading to overfitting and loss of generalization.  
Augmentation methods are often used to overcome these challenges by generating more data for training. Under this setting, we compare the accuracy of ResNet-18 \cite{resnet} trained using only $10$ images per class of the original Flower102 dataset. 
As shown in Table \ref{tab:data_scaricty}, our method consistently outperforms other mixup methods in situations where data is limited yielding accuracies of \numbersBlue{$77.14\%$} Top-1 and \numbersBlue{$74.12\%$} Top-5. Furthermore, our approach is designed to enhance the diversification of the training dataset. By leveraging our bespoke conditional prompts, \textsc{DiffuseMix}
artificially expands and enriches the overall data landscape, enabling a more robust learning of neural networks.

\begin{table}[t]
\renewcommand{\arraystretch}{1.1}
\setlength{\tabcolsep}{17pt}
\caption{Top-1 (\%) accuracy on \textbf{\emph{data scarcity}} task of ResNet-18 on Flower102 dataset where only 10 random images per class are used. Extended comparisons are provided in \textcolor{customviolet}{Appendix 7 Table 15}.}
\vspace{-7pt}
\resizebox{0.98\columnwidth}{!}{
\begin{tabular}{lcc}
\hline\thickhline
\rowcolor{mygray}
Method  & Valid  & Test  \\
\midrule
Vanilla\textsubscript{ \textcolor{blue}{(CVPR'16)}} \cite{resnet} & 64.48
 & 59.14  \\ 
SnapMix\textsubscript{ \textcolor{blue}{(AAAI'21)}} \cite{huang2021snapmix} &  65.71  & 59.79 \\
PuzzleMix\textsubscript{ \textcolor{blue}{(ICML'20)}} \cite{kim2020puzzle} & 71.56  & 66.71 \\
Co-Mixup\textsubscript{ \textcolor{blue}{(ICLR'20)}} \cite{kim2020co} & 68.17 & 63.20 \\
GuidedMixup\textsubscript{ \textcolor{blue}{(AAAI'23)}} \cite{kang2023guidedmixup}  & 74.74 & 70.44
\\ 
\hline
 \textbf{\textsc{DiffuseMix}}   & \textbf{77.14}  & \textbf{74.12} \\
\bottomrule
\end{tabular}
}
\vspace{-0.99em}

\label{tab:data_scaricty}
\end{table}

\subsection{Analysis and Discussions}

In this section, we provide a detailed ablation study and further analysis of various design choices in \textsc{DiffuseMix}.

\begin{table}[b]
\centering
\caption{Ablation study using Stanford Cars (cars) and Flowers102 (Flow) datasets. Top-1 and Top-5 accuracies are reported with \emph{\textbf{different combinations}} of \(I_{i}\): Input image, \(\hat{I}_{ij}\): Generated images using prompts \(p_j\), \(H_{iju}\): Hybrid images using random mask \(M_u\), and \(F_{v}\): fractal images used to obtain final blended image \(A_{ijuv}\).}
\scalebox{0.85}{
\begin{tabular}{cl|ccccccc}
& \(I_i\) & \ding{51} & \ding{51}   & -  & - & -  & -  \\
& \(\hat{I}_{ij}\)  & - & -   & \ding{51}  &\ding{51}  & -  & - \\
& \(H_{iju}\)  & - &- & -  & -  & \ding{51} & \ding{51}  \\
& \(F_{v}\) & - &\ding{51} & - &\ding{51} & - & \ding{51} \\ \hline
\multirow{2}{*}{\rotatebox{90}{Cars}} 
& Top-1  & 85.52 & 86.73   & 87.63 & 89.42 & 90.59 & \textbf{91.26}  \\
& Top-5 & 90.34 & 92.38 & 90.23 & 91.57 & 96.73 & \textbf{99.96}  \\  \hline
\multirow{2}{*}{\rotatebox{90}{Flow}} 
& Top-1  & 78.73 & 78.34   & 77.38 & 77.81 & 79.22 & \textbf{80.20}  \\
& Top-5 & 94.38 & 94.91 & 93.15 & 93.24 & 94.38  & \textbf{95.40}  \\
\end{tabular}
}
\label{tab:ablation}
\end{table}

\vspace{5pt}
\noindent \textbf{Ablation Studies:} 
To evaluate the importance of each component we conduct an ablation study by removing each component and report the observed performance on ResNet-50 using Stanford Cars and Flowers-102 datasets in Table \ref{tab:ablation}. 

When all of the components of \textsc{DiffuseMix} are removed, the baseline (vanilla) using only original images $I_i$ obtains Top-1 and Top-5 accuracies of \numbersBlue{$85.52\%$}  \& \numbersBlue{$90.34\%$} on Cars dataset and \numbersBlue{$78.83\%$}  \& \numbersBlue{ $94.38\%$} on Flowers dataset.
Next, we add fractal blending ($F_v$) to the input images resulting in slight performance gains on both datasets.
Further, we remove both concatenation ($H_{iju}$) and fractal blending ($F_v$) by conducting experiments using generated images ($\hat{I}_ij$) directly as augmented images to train the network. This setting brings the method closer to the approaches proposed in \cite{trabucco2023effective,azizi2023synthetic} which utilize the images generated using diffusion models directly as augmented images.
In this experiment, the accuracies obtained on Cars dataset are \numbersBlue{$87.63\%$} Top-$1$ and \numbersBlue{$90.23\%$} Top-5. Similarly, we observe the accuracies of 
\numbersBlue{$77.38\%$}  and \numbersBlue{$93.15\%$} on Flowers$102$ dataset. The results are consistent with the findings in \cite{azizi2023synthetic} that direct use of generated images may not yield significant performance gains over the vanilla method.
Further, we carry out experiments by using generated images with fractal blending to train the model. On the Cars dataset, we observe slight performance gains over the vanilla model with accuracies of \numbersBlue{$89.42\%$}  Top-1 and \numbersBlue{$91.57\%$} Top-5. However, in Flowers dataset, we observe a slight performance degradation. This demonstrates that to unleash the maximum benefits of fractal blending, it should be accompanied by more diversity in the training dataset. 
Next, we just remove the fractal blending from \our while incorporating concatenation between generated ($\hat{I}_ij$) and original ($I_i$) images to create hybrid images ($H_{iju}$) used as augmented training examples. This setting increases the accuracies to \numbersBlue{$90.59\%$} Top-1 \& \numbersBlue{$96.73\%$}  Top-5 on Cars dataset and 
\numbersBlue{$79.22\%$} Top-1 \& \numbersBlue{$94.38\%$} Top-5 on Flowers dataset.  These performance improvements are due to the availability of both generated and original image contents in each augmented image, highlighting the importance of the concatenation step in \textsc{DiffuseMix}.
Finally, when all components including fractal blending are used, the best accuracies of \numbersBlue{$91.26\%$} Top-1 \& \numbersBlue{$99.96\%$} Top-5 on Cars dataset and \numbersBlue{$80.20\%$} Top-1 \& \numbersBlue{$95.40\%$} Top-5 on Flowers dataset are achieved. Overall, consistent gains achieved with each added component signifies the design choices in \textsc{DiffuseMix} towards an effective image augmentation technique.



\begin{table}[t]
\setlength{\tabcolsep}{10pt}
\renewcommand{\arraystretch}{1.05}

\caption{Top-1 (\%) accuracy of \textsc{DiffuseMix} on \emph{\textbf{fine-tuning}}  experiments using ImageNet pretrained ResNet-50.}
\vspace{-7pt}
\centering
\resizebox{0.97\columnwidth}{!}{
\begin{tabular}{lccc}
\hline\thickhline
\rowcolor{mygray}
 Method  & Flower102  & Aircraft  & Cars \\ 
\midrule
Vanilla\textsubscript{\textcolor{blue}{(CVPR'16)}}  \cite{resnet} & 94.98 & 81.60 & 88.08 \\
AA\textsubscript{\textcolor{blue}{(CVPR'19)}}  \cite{autoaugment} & 93.88 & 83.39 & 90.82 \\
RA\textsubscript{\textcolor{blue}{(NIPS'20)}}  \cite{cubuk2020randaugment} & 95.23 &  82.98 & 89.28 \\
Fast AA\textsubscript{\textcolor{blue}{(NIPS'19)}}  \cite{lim2019fast} & 96.08 & 82.56 & 89.71 \\
AdaAug\textsubscript{\textcolor{blue}{(ICLR'22)}}  \cite{cheung2021adaaug} & 97.19  & 83.97  & 91.18  \\ 
\hline
\textbf{\textsc{DiffuseMix}}  & \textbf{98.02}  &  \textbf{85.65} &  \textbf{93.17} \\
\bottomrule
\end{tabular}
}
\vspace{-0.99em}
\label{tab:fine_tunning}
\end{table}

\noindent\textbf{Number of Prompts Vs. Performance:} 
The variety of augmented images is determined by the number of random prompts which is an important factor in \textsc{DiffuseMix}. A higher number of prompts corresponds to more diverse training samples. Figure \ref{fig:numofprompts} shows the effect of increasing the number of prompts on the performance using three different datasets including Birds, Aircraft and Cars. Increasing the number of prompts consistently yields performance gains on all three datasets using both Top-1 and Top-5 accuracy. However, the peak performances are achieved when all ten prompts proposed in our approach are used for training. 
In almost all datasets, the increasing accuracy trends can be seen even at $10$ prompts which shows that the addition of more prompts may further improve the performance. However, it will also incur more computational costs. 

\noindent\textbf{Increasing Masks Vs Performance:} We conduct experiments using Oxford Flower$102$ dataset to study the impact of various masks on the overall performance of \our and report the results in Table \ref{tab:mask_ablation}. Using even only one kind of mask, vertical in this example, our approach achieves 
significantly higher accuracies than the vanilla baseline.
When both vertical and horizontal masks are used, the accuracies improve further. 
However, the best accuracies 
are achieved when both horizontal and vertical masks are used along with random flipping between the positions of input and generated images adding more diversity to the training data. 
It is also possible to use the masking techniques from previous approaches, such as \cite{cutmix, lee2020smoothmix}, in \textsc{DiffusMix}. We provide additional analysis on this in \textcolor{customviolet}{Appendix 4.}






\par

\begin{table}
\renewcommand{\arraystretch}{1.2}
\centering
\caption{Ablation on the \emph{\textbf{effects of masking}}  in \our on Flower102 dataset. All variants yield notably superior results compared to vanilla on ResNet-50. However, best results are achieved when all four vertical and horizontal masks are used.
}
\vspace{-7pt}
\scalebox{0.85}{
\begin{tabular}{lcc}
\hline\thickhline
\rowcolor{mygray} 
\multicolumn{1}{l}{Mask} & \multicolumn{1}{c}{Top-1 (\%)} & \multicolumn{1}{c}{Top-5 (\%)} \\

\hline
 Vanilla\textsubscript{\textcolor{blue}{(CVPR'16)}}  \cite{resnet} & 89.74 & 94.38 \\ 
 Ver Mask ( \begin{tikzpicture}[baseline=-0.125ex]
            \fill[black] (0,0) rectangle (0.125cm,0.25cm); 
            \draw (0,0) rectangle (0.25cm,0.25cm); 
          \end{tikzpicture}
 ) & 94.02 & 98.42 \\
 Hor + Ver Masks ( 
 \begin{tikzpicture}[baseline=-0.125ex]
            \fill[black] (0,0) rectangle (0.125cm,0.25cm); 
            \draw (0,0) rectangle (0.25cm,0.25cm); 
          \end{tikzpicture},
 
 \begin{tikzpicture}[baseline=-0.125ex]
            \fill[black] (0,0) rectangle (0.25cm,0.125cm); 
            \draw (0,0) rectangle (0.25cm,0.25cm); 
          \end{tikzpicture} ) & 94.27 & 99.03 \\
 Hor + Ver + Flipping ( \begin{tikzpicture}[baseline=-0.125ex]
            \fill[black] (0,0) rectangle (0.125cm,0.25cm); 
            \draw (0,0) rectangle (0.25cm,0.25cm); 
          \end{tikzpicture} ,
           \begin{tikzpicture}[baseline=-0.125ex]
    \fill[black] (0.125cm,0) rectangle (0.25cm,0.25cm); 
    \draw (0,0) rectangle (0.25cm,0.25cm); 
\end{tikzpicture},
          
          \begin{tikzpicture}[baseline=-0.125ex]
            \fill[black] (0,0) rectangle (0.25cm,0.125cm); 
            \draw (0,0) rectangle (0.25cm,0.25cm); 
          \end{tikzpicture},
          
          \begin{tikzpicture}[baseline=-0.125ex]
    \fill[black] (0,0.125cm) rectangle (0.25cm,0.25cm); 
    \draw (0,0) rectangle (0.25cm,0.25cm); 
\end{tikzpicture}
)  & 95.37 & 99.39 \\
\bottomrule
\end{tabular}
}
\vspace{-0.99em}
\label{tab:mask_ablation}
\end{table}

\section{Conclusion}
\label{sec:conclusion}
In this paper, we introduced \textsc{DiffuseMix}, a data augmentation technique based on diffusion models to increase diversity in the data while preserving the original semantics of the input image. It involves \emph{generation, concatenation, and fractal blending}  steps to create the final augmented image. On multiple tasks such as general classification, fine-grained classification, data scarcity, fine-tuning, and adversarial robustness involving several benchmark datasets including 
\emph{ImageNet-1k, Tiny-ImageNet-200, CIFAR-100, Oxford Flower102, Caltech Birds, Stanford-Cars, and FGVC Aircraft}, \our demonstrates consistent performance gains and outperforms existing SOTA image augmentation methods.\\
\noindent\textbf{Limitations:} 
\our has two notable limitations: (1) Image generation relies heavily on text prompts and a wrong textual input may lead to unrealistic results. We address this issue by proposing a set of \textit{filter-like} prompts generally applicable to a wide range of natural images. 
(2) \our requires additional overheads for generating images (more on this in \textcolor{customviolet}{Appendix 2}).
This is a small price to pay for a guaranteed better convergence of large-scale classification models and can be mitigated by generating and storing augmented images once. \\
\noindent\textbf{Acknowledgements:} We are thankful to Hamza Saleem for the fruitful insights. Arif Mahmood is funded by the Information Technology University of Punjab, Pakistan.



{
    \small
    \bibliographystyle{ieeenat_fullname}
    \bibliography{main}
}
\clearpage
\setcounter{section}{0}

\maketitlesupplementary

\definecolor{deemph}{gray}{0.6}
\newcommand{\gc}[1]{\textcolor{deemph}{#1}}
\newcommand{\Rows}[1]{\multirow{2}{*}{#1}}
\definecolor{customgreen}{RGB}{0, 153, 0} 
\newcommand{\gain}[1]{\textcolor{customgreen}{\textbf{#1}}}
\definecolor{lossred}{RGB}{255, 0, 0} 
\newcommand{\loss}[1]{\textcolor{lossred}{\textbf{#1}}}

\noindent\textbf{\Large{Overview}}

\vspace{5pt}
This supplementary document contains additional results and discussions. Summarily, Section \ref{supp:blending_ratio} provides analysis on varying values of $\lambda$ which defines the ratio of blending between a fractal image and a hybrid image in \our. Section \ref{sec:augmentaiton_overhead_supp} provides comparisons of augmentation overhead between \our and existing image augmentation strategies with respect to their generalization performances. Section \ref{supp:bad_prompts} discusses examples of poorly constructed prompts and their effects on image generation. Section \ref{supp:diffuseMix_SOTA_CutMix} provides experimental results of using different masking strategies of the state-of-the-art methods with \textsc{DiffuseMix}. Section \ref{supp:faster_converges} provides the convergence analysis of \textsc{DiffuseMix}. Section \ref{supp:visulization_of_pts} provides more visualizations of the augmented training images obtained using \our. Section \ref{lab:suppl_performance_Eval} provides a complete list of general and fined-grained results. Section \ref{supp:fractal_dataset} provides some visual examples of the collected fractal image dataset.

\section{Fractal Blending Ratio} 
\label{supp:blending_ratio}
We experiment and observe the effect of varying fractal blending ratio \(\lambda\)  in \textsc{DiffuseMix} and report Top-$1$ accuracy (\%) results on Flower102 dataset in Table \ref{tab:blending}. The values of \(\lambda\) are varied from $0.1$ to $0.5$. A higher value of \(\lambda\) indicates a stronger ratio of fractal image blending.
\par
The baseline, ResNet50 without any augmentation, yields a top-$1$ accuracy of 
\numbersBlue{$78.73\%$}. Compared to this, \our{}  yields consistent performance gains with all values of \(\lambda\). The best performance of \our{} is observed at \(\lambda=0.2\), where the top-1 accuracy peaks at \numbersBlue{$81.30\%$}. However, generally, the performance remains better with a reasonable value of \(\lambda\). It starts dropping when the value of \(\lambda\) becomes too high.
This suggests that higher fractal blending ratios may introduce too much complexity or noise into the original data, which adversely affects the model's performance.

\begin{figure}[t]
    \centering
    \includegraphics[width=\linewidth]{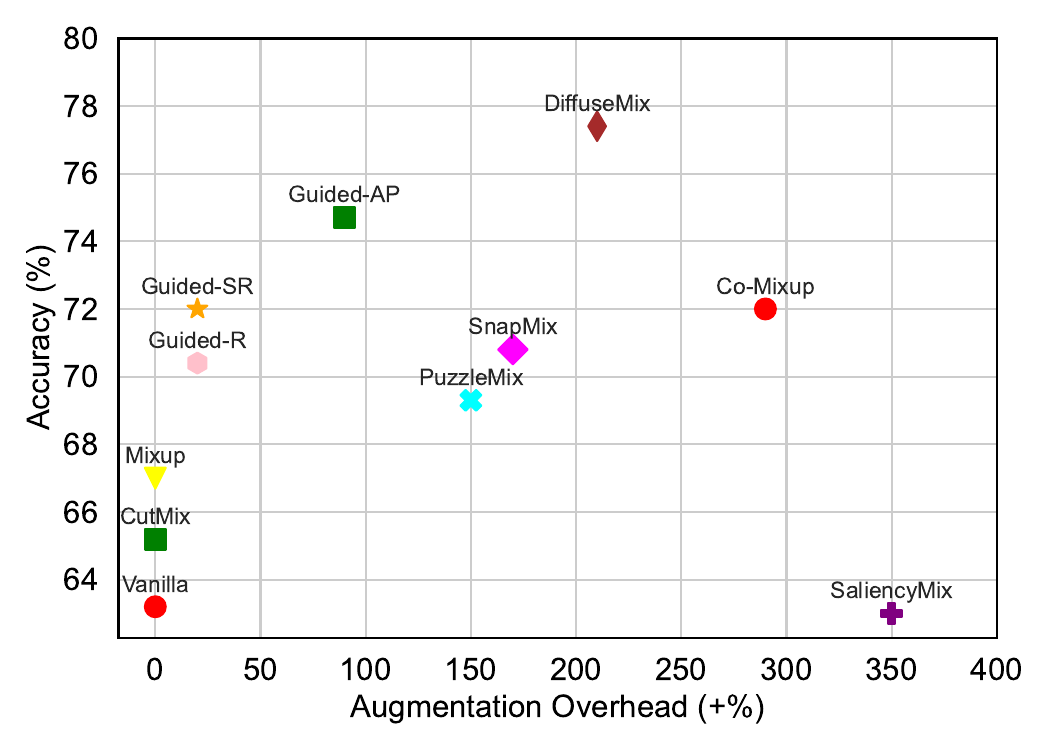}
    \caption{Augmentation overhead (+\%) - accuracy (\%) plot on CUB-200-2011 dataset with batch size $32$.}
    \label{fig:augOverhead}
\end{figure}

\begin{figure*}[t]
\centering
\begin{subfigure}{.19\textwidth}
  \centering
  \includegraphics[width=\linewidth]{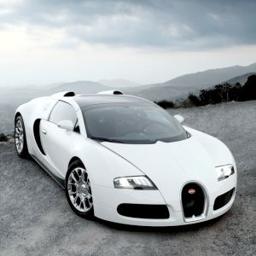}
  \caption{Stanford Cars}
\end{subfigure} \hspace{3pt}
\begin{subfigure}{.19\textwidth}
  \centering
  \includegraphics[width=\linewidth]{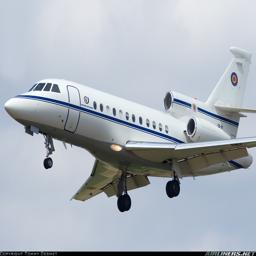}
  \caption{FGVC Aircraft}
\end{subfigure}\hspace{4pt}
\begin{subfigure}{.19\textwidth}
  \centering
  \includegraphics[width=\linewidth]{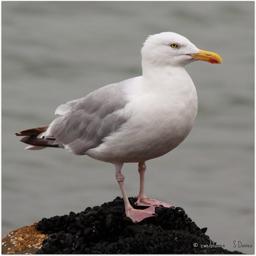}
  \caption{CUB-200-2011}
\end{subfigure}\hspace{4pt}%
\begin{subfigure}{.19\textwidth}
  \centering
  \includegraphics[width=\linewidth]{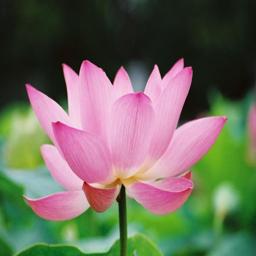}
  \caption{Oxford 102 Flower}
\end{subfigure}\hspace{4pt}%
\begin{subfigure}{.19\textwidth}
  \centering
  \includegraphics[width=\linewidth]{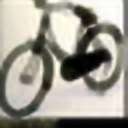}
  \caption{CIFAR100}
\end{subfigure}\hspace{4pt}%

\begin{subfigure}{.19\textwidth}
  \centering
  \includegraphics[width=\linewidth]{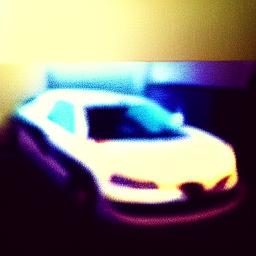}
  \caption*{ \textbf{\color{blue}{\emph{Sunset photo that looks like it's taken with 1990s camera}}}}
\end{subfigure} \hspace{3pt}
\begin{subfigure}{.19\textwidth}
  \centering
  \includegraphics[width=\linewidth]{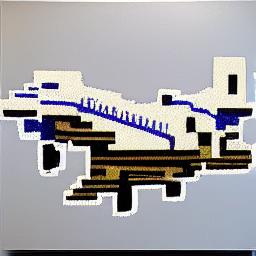}
  \caption*{ \textbf{\color{blue}{\emph{A painting that is too small for its body}}}}
\end{subfigure}\hspace{4pt}
\begin{subfigure}{.19\textwidth}
  \centering
  \includegraphics[width=\linewidth]{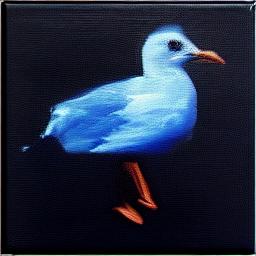}
  \caption*{ \textbf{\color{blue}{\emph{A cityscape that is too dark to see anything}}}}
\end{subfigure}\hspace{4pt}%
\begin{subfigure}{.19\textwidth}
  \centering
  \includegraphics[width=\linewidth]{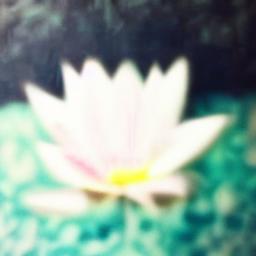}
  \caption*{ \textbf{\color{blue}{\emph{Photograph that is washed it looks like it is made of paper}}}}
\end{subfigure}\hspace{4pt}%
\begin{subfigure}{.19\textwidth}
  \centering
  \includegraphics[width=\linewidth]{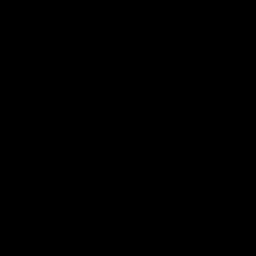}
  \caption*{ \textbf{\color{blue}{\emph{A distorted, warped painting of a landscape}}}}
\end{subfigure}\hspace{4pt}%
\caption{\textbf{First row:} original training image samples from different datasets \emph{such as} Oxford-102 Flower \cite{nilsback2008automated}, Stanford Cars \cite{krause20133d}, and Aircraft~\cite{maji2013fine},  CUB-200-2011, and CIFAR100. \textbf{Second row:} Corresponding generated images show that the usage of descriptive prompts (blue text) results in poor images not feasible for training. When generating images on the CIFAR100 dataset, several additional challenges may occur due to the small size of the images. For example, the image in the last column taken from CIFAR-100 with its corresponding prompt results in a \textbf{\emph{{black}}} image containing no visible output.}
\label{fig:poor_prompts}
\end{figure*}

\section{Augmentation Overhead}
\label{sec:augmentaiton_overhead_supp}

We compare the computational overhead of several existing SOTA image augmentation methods and \our{} with respect to the performance gains. Following Kang \etal \cite{kang2023guidedmixup}, we define the augmentation overhead $\mathcal{A_{O}}$ as:
$$\mathcal{A_{O}}=\frac{\mathcal{T}_{aug}-\mathcal{T}_{van}}{\mathcal{T}_{van}}\times 100 (\%),$$
where $\mathcal{T}$ is the total image generation and training time, and $\mathcal{T}_{van}$ is the training time of the baseline network without any augmentation. Although image generation process in \our{} can be expedited by using parallel-processing, We do not utilize it to provide a fair comparison. It can be seen in Figure \ref{fig:augOverhead} that \our{} provides a good tradeoff between performance and augmentation overhead by outperforming all existing approaches in terms of accuracy while providing significantly lower augmentation overhead compared to Co-Mixup and SaliencyMix approaches. 
Moreover, \our{} can also be optimized further by saving the generated images offline once before carrying out any number of subsequent trainings. This may make it significantly faster to perform several experiments on a training model, particularly for optimization and research purposes.

\begin{table} [t]
\centering
\caption{Impact of varying fractal blending ratio in \textsc{DiffuseMix}. Top-1 accuracy is reported using ResNet-50 on Flower102 dataset.}
\begin{tabular}{lcc}
\hline\thickhline
\rowcolor{mygray}
Methods & Top-1 (\%)  \\
\hline
ResNet$50$\textsubscript{\textcolor{blue}{(CVPR'16)}} \cite{resnet}  & 78.73  \\

\hspace{0.2mm} + \emph{  DiffuseMix}  ($\lambda=0.1$) & 79.81  \\
\hspace{0.2mm} + \emph{DiffuseMix}  ($\lambda=0.2$) & \textbf{81.30}  \\
\hspace{0.2mm} + \emph{DiffuseMix}  ($\lambda=0.3$) & 80.97  \\
\hspace{0.2mm} + \emph{DiffuseMix}  ($\lambda=0.4$) & 79.16  \\
\hspace{0.2mm} + \emph{DiffuseMix}  ($\lambda=0.5$) & 78.97  \\
\bottomrule
\end{tabular}
\label{tab:blending}
\end{table}

\section{Prompt Selection}
\label{supp:bad_prompts}


Diffusion Models rely heavily on prompts \cite{du2023stable}.
Therefore, the intuition behind designing our \emph{bespoke conditional prompts} is to introduce the type of prompts that may edit the image in a way that preserves structural information and can easily be applied to a range of diverse datasets. 
To this end, as described in the manuscript, we propose to use \textit{filter-like} prompts such as snowy, sunset, rainbow, etc. and demonstrate their effectiveness in training robust classifiers. 

Conversely, in this section, we discuss \textit{bad} prompts that may not be a good fit for the image generation step of \our{}.
Some examples of such prompts are shown in Figure \ref{fig:poor_prompts}. 
More descriptive and overly complicated prompts generate images that may be too different from the original distribution. The resultant images contain unrealistic foregrounds and backgrounds, rendering these useless for the training of a classifier. This reiterates the importance of our proposed \textit{filter-like bespoke conditional prompts} that do not induce unwanted changes to the training images.

\begin{figure*}
    \centering
    \begin{subfigure}{.31\textwidth}
        \centering
        \includegraphics[width=\linewidth]{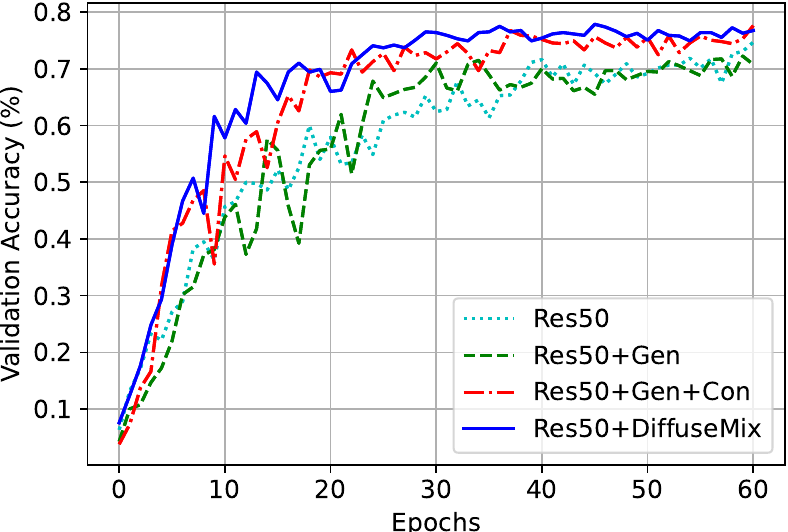}
        \caption{Top-$1$ Accuracy (\%)}
            \label{sub:top5_acc}
    \end{subfigure} 
    \begin{subfigure}{.31\textwidth}
        \centering
        \includegraphics[width=\linewidth]{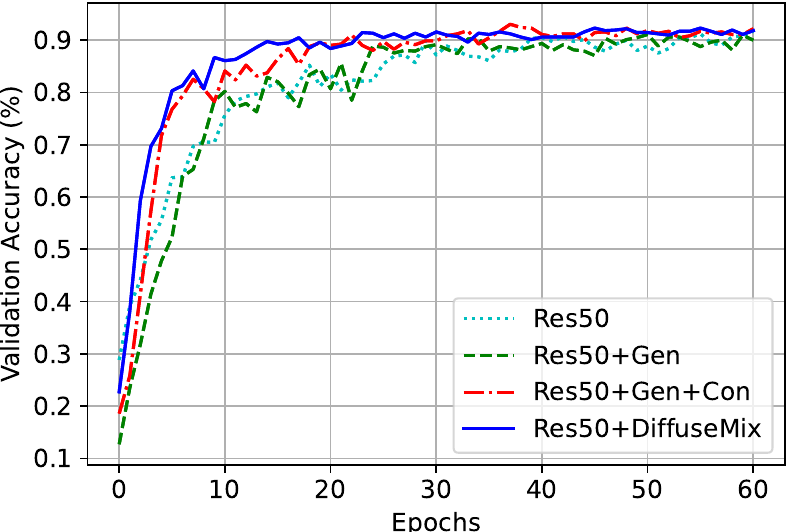}
        \caption{Top-$5$ Accuracy (\%)}
                \label{sub:top1_acc}
    \end{subfigure}
    \hspace{2mm} 
    \begin{minipage}[c]{2mm}
        \raisebox{0cm}[0pt][0pt]{ 
            \begin{tikzpicture}
                \draw[dotted, line width=1.5pt] (0,0) -- (0,-4cm); 
            \end{tikzpicture}
        }
    \end{minipage}
    \hspace{2mm} 
    \begin{subfigure}{.31\textwidth}
        \centering
        \includegraphics[width=\linewidth]{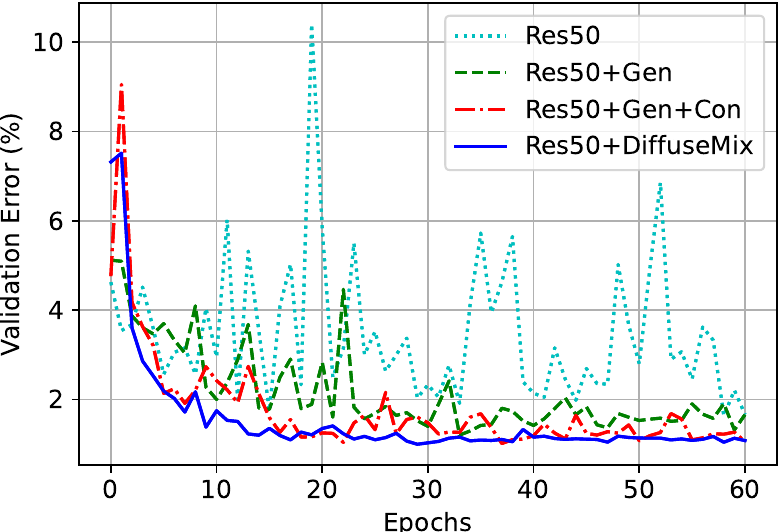}
        \caption{Validation Error (\%)}
        \label{sub:validation_error}
    \end{subfigure}
    \caption{\emph{\textbf{On left side:}} The curves of top-$1$ and top-$5$ accuracy show an increasing trend during initial $60$ epochs and remain stable towards the end on Flower102 dataset. This same behavior can also be seen in top-$5$ accuracy. |our{} enables smoother training and better convergence while avoiding overfitting. \emph{\textbf{On right side:}}  Similar to the accuracy plots, using \our{} demonstrates a smoother decrease in validation error compared to ResNet$50$ or other variants. \textit{Best viewed in color.}}
    \label{fig:loss_acc}
\end{figure*}

\begin{table}[!b]
\centering
\caption{Combining \our{} with SOTA image augmentation methods by replacing the image concatenation technique of \our{} with the masking techniques proposed in \cite{cutmix} \& \cite{mixup}. While \our{} provides consistent gains in these settings, the best performance of \numbersBlue{$81.30\%$} is achieved when our originally proposed method is used.}

\begin{tabular}{lc}
\hline\thickhline
\rowcolor{mygray} 
\multicolumn{1}{l}{Method} & \multicolumn{1}{c}{Top-1 (\%)}  \\
\hline
ResNet$50$\textsubscript{\textcolor{blue}{(CVPR'16)}} \cite{resnet}  & 78.73   \\
\hspace{0.2mm} + CutMix \cite{cutmix} & 79.22  \\
\hspace{0.2mm} + CutMix \cite{cutmix} + \textsc{DiffuseMix}  & 79.58  \\
\hspace{0.2mm} + Mixup \cite{mixup} & 79.34  \\
\hspace{0.2mm} + Mixup \cite{mixup} + \textsc{DiffuseMix} & {80.20}   \\
\hspace{0.2mm} + \textsc{DiffuseMix} & \textbf{81.30}   \\
\bottomrule
\end{tabular}
\label{tab:diffuseMix_Combine_SOTA}
\end{table}

\section{\textsc{DiffuseMix} with SOTA Methods}
\label{supp:diffuseMix_SOTA_CutMix}
In a series of experiments, we combine \our{} with existing image augmentation approaches \cite{cutmix,mixup} to see if any performance gain is observed. Particularly, We replace our masking approach with the masking used in the existing methods while retaining the rest of the pipeline of \our{} same. 

For CutMix + \our{}, we replace the concatenation step of \our{} with the random cropping of CutMix. To this end, we randomly crop a patch from the generated image and paste it onto the original image whereas the other stages remain the same. For Mixup + \textsc{DiffuseMix}, we replace concatenation with the pixel blending of original and generated images as proposed in \cite{mixup} while the rest of the steps remain intact.
The results are summarized in Table \ref{tab:diffuseMix_Combine_SOTA}.
Using CutMix \cite{cutmix} or Mixup \cite{mixup} methods yields improvements over baseline ResNet50 training. However, when our proposed approach is added to the existing methods, further performance gains are observed.
Top performance is finally observed with our \our{}, which demonstrates the importance of forming hybrid images by concatenating original and generated images.

\section{\textsc{DiffuseMix} Convergence}
\label{supp:faster_converges}

\par
\vspace{5pt}
\noindent \textbf{Analysis on Top-1 and Top-5 Accuracy:} In a series of experiments, we carry out an ablation to observe the top-1 and top-5 accuracies of \our{} and its variants formed by removing the components \emph{ (generation, concatenation, and fractal blending)} one by one.

As seen in Figures \ref{sub:top5_acc} and  \ref{sub:top1_acc}, the \textsc{Res50+DiffuseMix} demonstrates generally better performance with convergence at \numbersBlue{$77.26\%$} accuracy, closely followed by \textsc{DiffuseMix+Gen+Con} at \numbersBlue{$75.79\%$}, and Res50 at \numbersBlue{$76.41\%$}. The \textsc{DiffuseMix+Gen} model performs significantly lower yielding \numbersBlue{$73.96\%$} accuracy.
As discussed in the manuscript Section 4, using generated images directly for the training may lead to deteriorated performance, which is re-validated in these experiments. This also shows the importance of each step proposed in \our{} towards robust training more robust classifiers. Overall, similar trends are observed in Top-5 accuracy results (Figure \ref{sub:top1_acc}).


\par
\vspace{5pt}
\noindent \textbf{Analysis on validation loss:} As seen in (Figure \ref{sub:validation_error}), it is clearly noticeable that \textsc{DiffuseMix} helps in model convergence and overall smooth decrease in validation loss during training. Res50 baseline shows a good start with lower initial loss. However, its loss starts fluctuating once the training is continued indicating a potential plateau in learning or its limitation in capturing more complex patterns. Compared to all variants, Res50+\textsc{DiffuseMix} benchmarks better convergence.  


\section{\textsc{DiffuseMix} Visualizations}
\label{supp:visulization_of_pts}

\begin{figure*}
\centering

\begin{subfigure}{.19\textwidth}
  \centering
  \includegraphics[width=\linewidth]{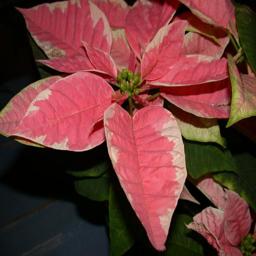}
  \caption{Poinsettia}
\end{subfigure} \hspace{3pt}
\begin{subfigure}{.19\textwidth}
  \centering
  \includegraphics[width=\linewidth]{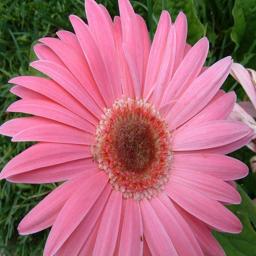}
  \caption{Barbeton Daisy}
\end{subfigure}\hspace{4pt}
\begin{subfigure}{.19\textwidth}
  \centering
  \includegraphics[width=\linewidth]{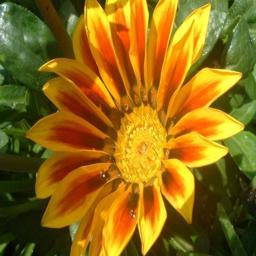}
  \caption{Gazania}
\end{subfigure}\hspace{4pt}%
\begin{subfigure}{.19\textwidth}
  \centering
  \includegraphics[width=\linewidth]{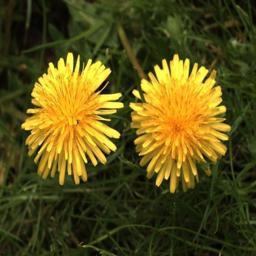}
  \caption{Dandelion}
\end{subfigure}\hspace{4pt}%
\begin{subfigure}{.19\textwidth}
  \centering
  \includegraphics[width=\linewidth]{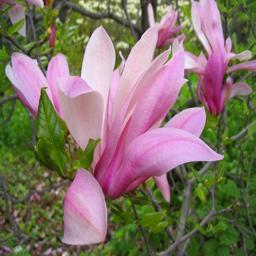}
  \caption{Magnolia}
\end{subfigure}\hspace{4pt}%

\begin{subfigure}{.19\textwidth}
  \centering
  \includegraphics[width=\linewidth]{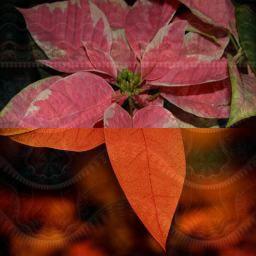}
  \caption{Poinsettia Autumn}
\end{subfigure} \hspace{3pt}
\begin{subfigure}{.19\textwidth}
  \centering
  \includegraphics[width=\linewidth]{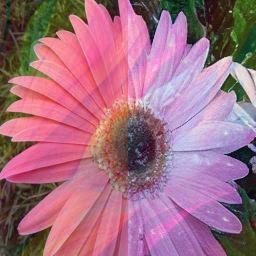}
  \caption{Barbeton Daisy Snowy}
\end{subfigure}\hspace{4pt}
\begin{subfigure}{.19\textwidth}
  \centering
  \includegraphics[width=\linewidth]{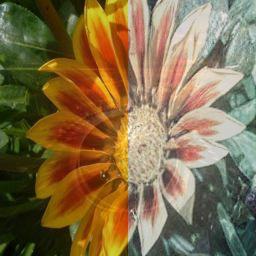}
  \caption{Gazania Crayon Sketch}
\end{subfigure}\hspace{4pt}%
\begin{subfigure}{.19\textwidth}
  \centering
  \includegraphics[width=\linewidth]{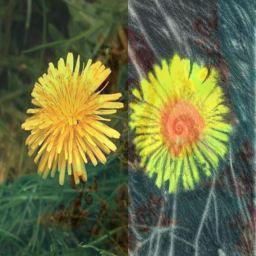}
  \caption{Dandelion Crayon Sketch}
\end{subfigure}\hspace{4pt}%
\begin{subfigure}{.19\textwidth}
  \centering
  \includegraphics[width=\linewidth]{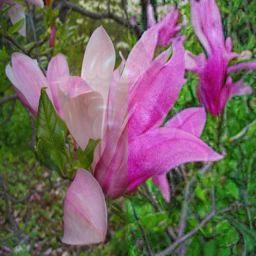}
  \caption{Magnolia Crayon Sketch}
\end{subfigure}\hspace{4pt}%
\caption{Illustration of original training images and \textsc{DiffuseMix} augmented images from the Oxford Flower$102$ dataset. {\textbf{First row:} showcases original, unaltered images of various flowers, including \emph{poinsettia, barbeton daisy, gazania, dandelion,} and \emph{Magnolia} classes. \textbf{Second row:} illustrates the transformative effects of the \textsc{DiffuseMix} augmentation method.
The effects of our custom-tailored prompts-based generation are visible on the generated portion of each image.
Overall, \our{} results in a diverse array of images with sufficient structural complexity and diversity to train robust classifiers.}}
\label{fig:dmix_flower102}
\end{figure*}

\begin{figure*}[b]
\centering

\begin{subfigure}{.19\textwidth}
  \centering
  \includegraphics[width=\linewidth]{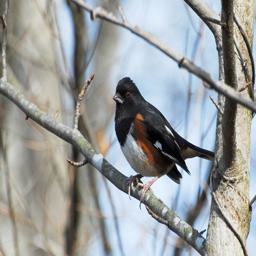}
  \caption{Eastern Towhee}
\end{subfigure}\hspace{4pt}%
\begin{subfigure}{.19\textwidth}
  \centering
  \includegraphics[width=\linewidth]{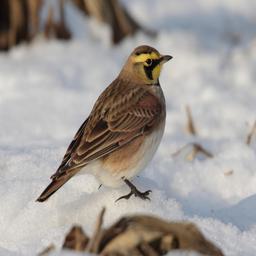}
  \caption{Horned Lark}
\end{subfigure}\hspace{4pt}%
\begin{subfigure}{.19\textwidth}
  \centering
  \includegraphics[width=\linewidth]{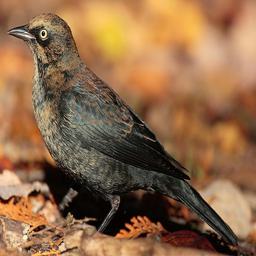}
  \caption{Rusty Blackbird}
\end{subfigure}\hspace{4pt}%
\begin{subfigure}{.19\textwidth}
  \centering
  \includegraphics[width=\linewidth]{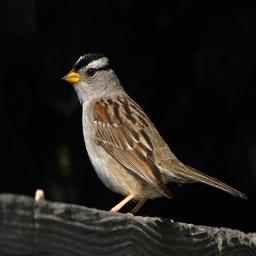}
  \caption{White Sparrow}
\end{subfigure}\hspace{4pt}%
\begin{subfigure}{.19\textwidth}
  \centering
  \includegraphics[width=\linewidth]{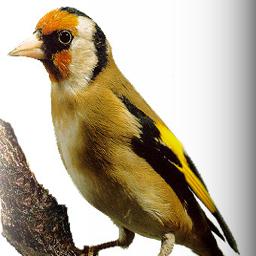}
  \caption{European Goldfinch}
\end{subfigure}\hspace{4pt}%

\begin{subfigure}{.19\textwidth}
  \centering
  \includegraphics[width=\linewidth]{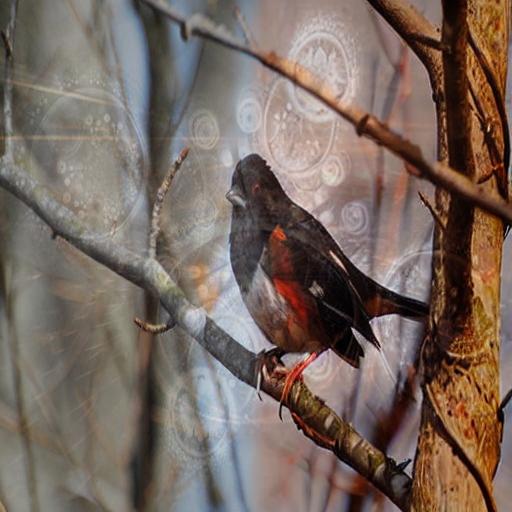}
  \caption{Eastern Towhee Sunset}
\end{subfigure}\hspace{4pt}%
\begin{subfigure}{.19\textwidth}
  \centering
  \includegraphics[width=\linewidth]{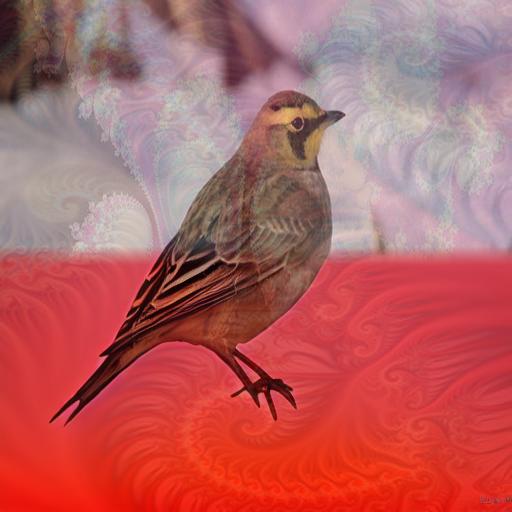}
  \caption{Horned Lark Sunset}
\end{subfigure}\hspace{4pt}%
\begin{subfigure}{.19\textwidth}
  \centering
  \includegraphics[width=\linewidth]{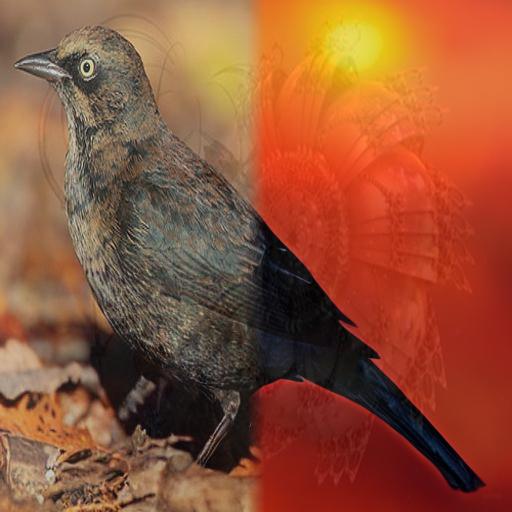}
  \caption{Rusty Blackbird Autumn}
\end{subfigure}\hspace{4pt}%
\begin{subfigure}{.19\textwidth}
  \centering
  \includegraphics[width=\linewidth]{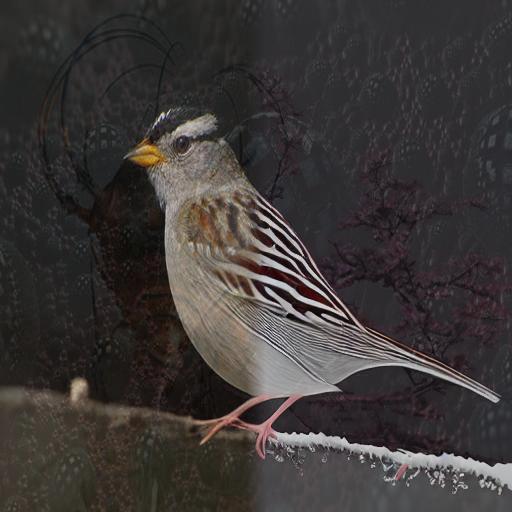}
  \caption{White Sparrow Snowy}
\end{subfigure}\hspace{4pt}%
\begin{subfigure}{.19\textwidth}
  \centering
  \includegraphics[width=\linewidth]{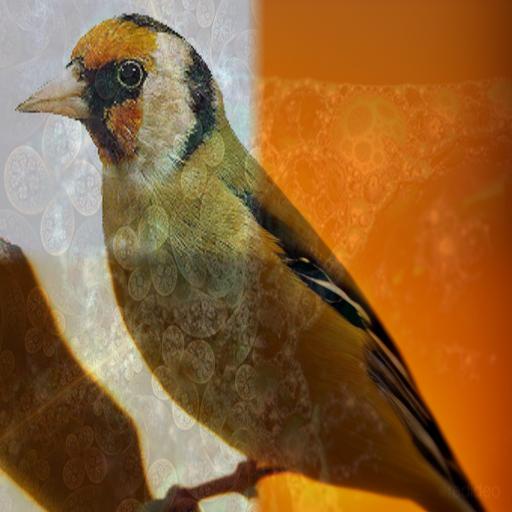}
  \caption{Europe Goldfinch Snowy}
\end{subfigure}\hspace{4pt}%

\caption{Original and \textsc{DiffuseMix}  augmented bird images from the Caltech-UCSD Birds-200-2011 dataset. \textbf{Top row:} displays a selection of original, high-resolution bird images, capturing the natural beauty and diversity of species such as the \emph{eastern towhee, horned lark, rusty blackbird, white sparrow,} and \emph{european goldfinch}. \textbf{Bottom row:} demonstrates the augmented images obtained using \textsc{DiffuseMix}. The augmented images are visually striking and contextually varied representations of the original subjects. 
}
\label{fig:curbirds_dmix}
\end{figure*}

\begin{figure*}[t]
\centering
\begin{subfigure}{.19\textwidth}
  \centering
  \includegraphics[width=\linewidth]{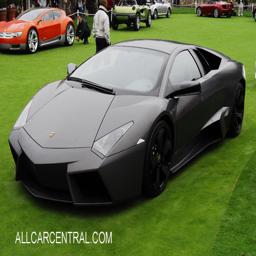}
  \caption{Lamborghini}
\end{subfigure}\hspace{4pt}%
\begin{subfigure}{.19\textwidth}
  \centering
  \includegraphics[width=\linewidth]{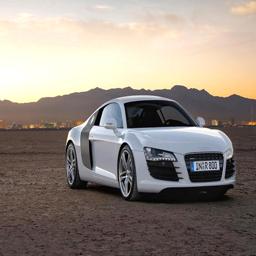}
  \caption{Audi R8}
\end{subfigure}\hspace{4pt}%
\begin{subfigure}{.19\textwidth}
  \centering
  \includegraphics[width=\linewidth]{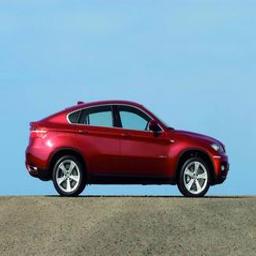}
  \caption{Bentley}
\end{subfigure}\hspace{4pt}%
\begin{subfigure}{.19\textwidth}
  \centering
  \includegraphics[width=\linewidth]{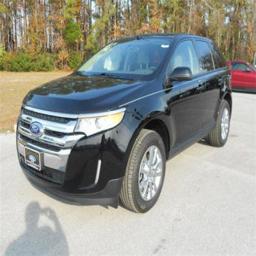}
  \caption{Ford Edge}
\end{subfigure}\hspace{4pt}%
\begin{subfigure}{.19\textwidth}
  \centering
  \includegraphics[width=\linewidth]{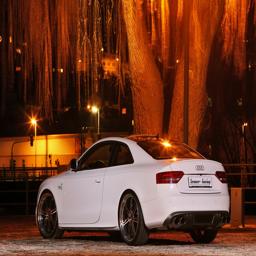}
  \caption{Audi S5}
\end{subfigure}\hspace{4pt}%

\begin{subfigure}{.19\textwidth}
  \centering
  \includegraphics[width=\linewidth]{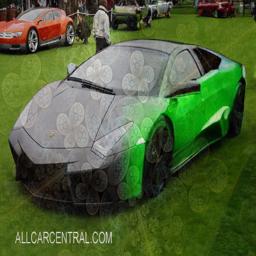}
  \caption{Lamborghini Aurora}
\end{subfigure}\hspace{4pt}%
\begin{subfigure}{.19\textwidth}
  \centering
  \includegraphics[width=\linewidth]{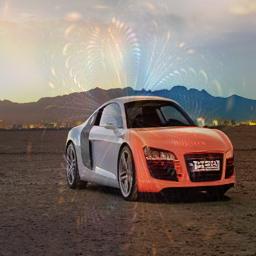}
  \caption{Audi R8 Rainbow}
\end{subfigure}\hspace{4pt}%
\begin{subfigure}{.19\textwidth}
  \centering
  \includegraphics[width=\linewidth]{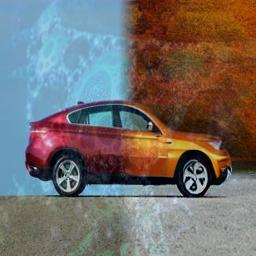}
  \caption{Bentley Autumn}
\end{subfigure}\hspace{4pt}%
\begin{subfigure}{.19\textwidth}
  \centering
  \includegraphics[width=\linewidth]{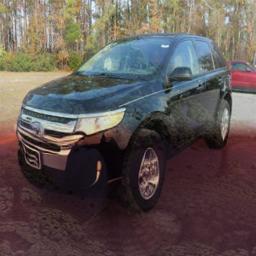}
  \caption{Ford Edge Sunset}
\end{subfigure}\hspace{4pt}%
\begin{subfigure}{.19\textwidth}
  \centering
  \includegraphics[width=\linewidth]{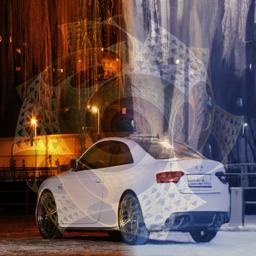}
  \caption{Audi S5 Snowy}
\end{subfigure}\hspace{4pt}%
\caption{\textbf{First row:} showcases original images from the Stanford Cars benchmark dataset, featuring unaltered depictions of various car models including a \emph{lamborghini, audi R8, bentley, ford edge} and \emph{audi S5}. \textbf{Second row:} presents the images transformed using our \textsc{DiffuseMix} method. The effects of prompts are visible in the generated portions of the images. For example, \emph{lamborghini} is changed to green when  \emph{aurora} prompt is applied, creating a vibrant image. The front side of \emph{audi R8} becomes more color-rich when it is generated with \emph{rainbow} prompt. The ambiance (background context) of \emph{bentley} transforms significantly when \emph{autumn} prompt is used. Similar diverse transformations are observed in other examples. These augmented images demonstrate the capability of \textsc{DiffuseMix} in generating visually enriched augmented images for better generalization.}
\label{fig:stan_cars_dMix}
\end{figure*}

\begin{figure*}[t]
\centering

\begin{subfigure}{.19\textwidth}
  \centering
  \includegraphics[width=\linewidth]{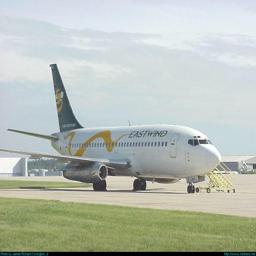}
  \caption{737-200}
\end{subfigure}\hspace{4pt}%
\begin{subfigure}{.19\textwidth}
  \centering
  \includegraphics[width=\linewidth]{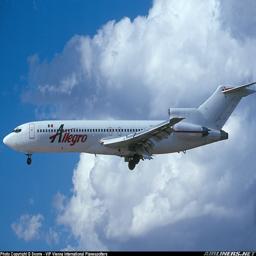}
  \caption{727-200}
\end{subfigure}\hspace{4pt}%
\begin{subfigure}{.19\textwidth}
  \centering
  \includegraphics[width=\linewidth]{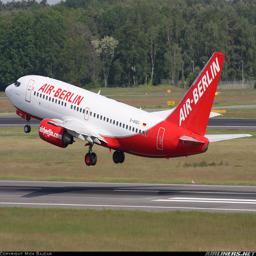}
  \caption{737-700}
\end{subfigure}\hspace{4pt}%
\begin{subfigure}{.19\textwidth}
  \centering
  \includegraphics[width=\linewidth]{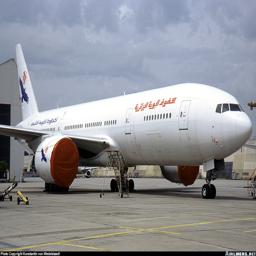}
  \caption{777-200}
\end{subfigure}\hspace{4pt}%
\begin{subfigure}{.19\textwidth}
  \centering
  \includegraphics[width=\linewidth]{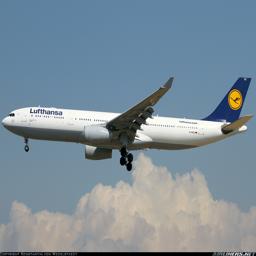}
  \caption{A330-300}
\end{subfigure}\hspace{4pt}%

\begin{subfigure}{.19\textwidth}
  \centering
  \includegraphics[width=\linewidth]{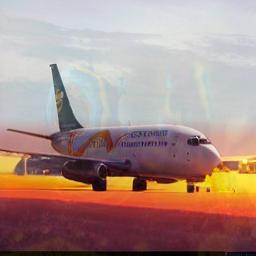}
  \caption{737-200 Sunset}
\end{subfigure}\hspace{4pt}%
\begin{subfigure}{.19\textwidth}
  \centering
  \includegraphics[width=\linewidth]{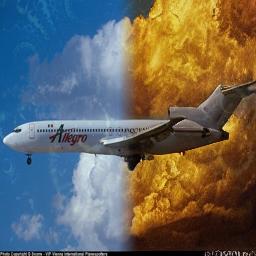}
  \caption{727-200 Autumn}
\end{subfigure}\hspace{4pt}%
\begin{subfigure}{.19\textwidth}
  \centering
  \includegraphics[width=\linewidth]{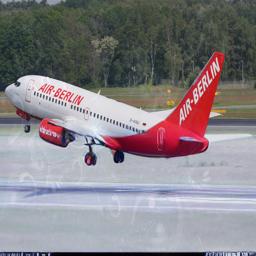}
  \caption{737-700 Snowy}
\end{subfigure}\hspace{4pt}%
\begin{subfigure}{.19\textwidth}
  \centering
  \includegraphics[width=\linewidth]{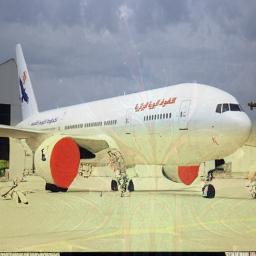}
  \caption{777-200 Ukiyo}
\end{subfigure}\hspace{4pt}%
\begin{subfigure}{.19\textwidth}
  \centering
  \includegraphics[width=\linewidth]{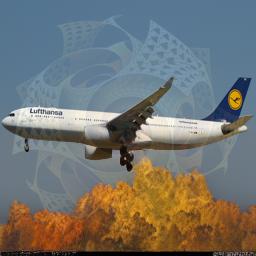}
  \caption{A330-300 Autumn}
  
\end{subfigure}\hspace{4pt}%
\caption{Illustration of original and 
\textsc{DiffuseMix} augmented Aircraft images from the FGVC-Aircraft benchmark dataset. \textbf{Top row:} presents original aircraft images, each portraying a distinct airplane including the $737-200$, $727-200$, $737-700$, $777-200$, and A$330-300$. These images highlight the design resemblance of various aircraft models, serving as a challenging resource for aircraft fined-grained image classification studies. \textbf{Bottom row:} showcases the augmented images obtained using  \textsc{DiffuseMix} for each corresponding input image. As seen, \our{} reimagined each aircraft with unique prompts such as \emph{sunset, autumn, snowy} and \emph{ ukiyo} resulting in a rich visual appearance with diverse contexts. 
This also illustrates how image augmentation can be used to simulate different environmental and stylistic scenarios, potentially enhancing the robustness and versatility of the dataset for training robust neural networks.}
\label{fig:dmix_aircraft}
\end{figure*}

In this section, we provide more visual examples of training images obtained using  \textsc{DiffuseMix}. 
As seen in Figure \ref{fig:dmix_flower102}, visualizing examples from Flower102 dataset, \textsc{DiffuseMix} enhances the overall variation of the images while retaining the interpretability of each example. For Caltech-UCSD Birds-200-2011 (Figure \ref{fig:curbirds_dmix}), compared to original images, the augmented images obtained using \textsc{DiffuseMix} exhibit greater clarity and diverse contexts. Similar visual features can be observed in Figure \ref{fig:stan_cars_dMix} and Figure \ref{fig:dmix_aircraft} showcasing examples from Cars and Aircraft datasets, respectively.


\vspace{10pt}

\begin{table}
\renewcommand{\arraystretch}{1.0}
\centering
\caption{Top-$1$ and Top-$5$ general classification accuracies comparison using PreActResNet-$18$. Compared numbers are taken either from the original papers or from \cite{kang2023guidedmixup}.}
\scalebox{0.85}{
\begin{tabular}{p{2.5cm}p{1.1cm}<{\centering}p{1.1cm}<{\centering}|p{1.1cm}<{\centering}p{1.2cm}<{\centering}}
\hline\thickhline
\rowcolor{mygray}
& \multicolumn{2}{c|}{\cellcolor{mygray}Tiny-ImageNet} & \multicolumn{2}{c}{\cellcolor{mygray}CIFAR-100} \\ \cline{2-5}
\rowcolor{mygray}
Method & \cellcolor{mygray}\begin{tabular}[c]{@{}c@{}}Top-1\\ Acc (\%) \end{tabular} & \cellcolor{mygray}\begin{tabular}[c]{@{}c@{}}Top-5\\ Acc (\%) \end{tabular} & \cellcolor{mygray}\begin{tabular}[c]{@{}c@{}}Top-1\\ Acc (\%) \end{tabular} & \cellcolor{mygray}\begin{tabular}[c]{@{}c@{}}Top-5\\ Acc (\%)\end{tabular} \\ 
\hline  
Vanilla \cite{resnet} & 57.23 & 73.65 & 76.33 & 91.02 \\
Mixup \cite{mixup} & 56.59 & 73.02 & 76.84 & 92.42 \\
Manifold \cite{verma2019manifold} & 58.01 & 74.12 & 79.02 & 93.37 \\
CutMix \cite{cutmix} & 56.67 & 75.52 & 76.80 & 91.91 \\
AugMix \cite{hendrycksaugmix} & 55.97 & 74.68 & 75.31 & 91.62 \\
PixMix \cite{hendrycks2022pixmix} & - & - & 79.70 & -  \\
SaliencyMix \cite{uddin2020saliencymix} & 56.54 & 76.14 & 79.75 & 94.71 \\
Guided-SR \cite{kang2023guidedmixup} & 55.97 & 74.68 & 80.60 & 94.00 \\
PuzzleMix \cite{kim2020puzzle} & 63.48 & 75.52 & 80.38 & 94.15 \\
Co-Mixup \cite{kim2020co} & 64.15 & - & 80.15 & - \\
Guided-AP \cite{kang2023guidedmixup} & 64.63  & 82.49  & 81.20  & 94.88  \\

\hline
\textbf{\textsc{DiffuseMix}}  & \textbf{65.77} & \textbf{83.66} & \textbf{82.50}  & \textbf{95.41} \\
\hline
\end{tabular}
}
\label{tab:supp_CIFAR_Tiny}
\end{table}

\section{Performance Evaluation}
\label{lab:suppl_performance_Eval}
Extended versions of the performance tables are provided in this section.
\subsection{General Classification}

In Table \ref{tab:supp_CIFAR_Tiny}, Vanilla method serves as our baseline, achieving Top-1 accuracies of \numbersBlue{$57.23\%$} and \numbersBlue{$76.33\%$} on Tiny-ImageNet and CIFAR-100 respectively, setting a benchmark for subsequent comparisons. For Mixup, we observe a slight decline in performance on Tiny-ImageNet to \numbersBlue{$56.59\%$} Top-1 accuracy but a marginal improvement on CIFAR-100, reaching \numbersBlue{$76.84\%$}. Conversely, Manifold Mixup marks notable performance gains, especially on CIFAR-100 with a Top-1 accuracy of \numbersBlue{$79.02\%$}. CutMix slightly improves over the baseline on Tiny-ImageNet, whereas AugMix shows a decrement, particularly on CIFAR-100 with a \numbersBlue{$75.31\%$} Top-1 accuracy.
\par
PixMix introduces variations in the source image instead of mixing two input images. Compared to baseline, PixMix excels on CIFAR-100 with a \numbersBlue{$79.70\%$} Top-1 accuracy. SaliencyMix, which uses saliency to mix different portions of images, also shows promising results. Particularly on CIFAR-100, it achieves a Top-1 accuracy of \numbersBlue{$79.75\%$}. The Guided-SR method performs slightly lower compared to AugMix on Tiny-ImageNet but stands out on CIFAR-100 with \numbersBlue{$80.60\%$} Top-1 accuracy, indicating its effectiveness. PuzzleMix and Co-Mixup introduce more complex ways to augment data, with PuzzleMix reaching a notable \numbersBlue{$63.48\%$} Top-1 accuracy on Tiny-ImageNet. Co-Mixup tops these methods on Tiny-ImageNet with \numbersBlue{$64.15\%$} Top-1 accuracy but does not maintain this lead on CIFAR-100. Guided-AP pushes the performance boundaries further by achieving superior accuracies among its predecessors, e.g., \numbersBlue{$81.20\%$} Top-1 accuracy on CIFAR-100.
\par
\textsc{DiffuseMix}, our proposed method, which surpasses all prior techniques by securing the highest accuracies: \numbersBlue{$65.77\%$} Top-1 on Tiny-ImageNet and \numbersBlue{$82.50\%$} Top-1 on CIFAR-100. Our approach not only surpasses the conventional mixup strategies but also sets a new standard in enhancing the generalization of deep learning models. The performance of \textsc{DiffuseMix} stays consistent across the compared datasets, underlining its superior capability and efficiency.
\begin{table}
\centering
\caption{Top-1 and Top-5 accuracies comparison on ImageNet using ResNet-50. Compared numbers are taken either from the original papers or from \cite{kang2023guidedmixup}.}
\vspace{-7pt}
\scalebox{0.90}{
\begin{tabular}{p{4.0cm}p{1.3cm}<{\centering}p{1.3cm}<{\centering}}
\hline\thickhline
\rowcolor{mygray}
Method & Top-1 Acc.  & Top-5 Acc.  \\ 
\hline 
Vanilla\textsubscript{\textcolor{blue}{(CVPR'16)}} \cite{resnet} & 75.97 & 92.66 
\\
AugMix \textsubscript{\textcolor{blue}{(ICLR'20)}} \cite{hendrycksaugmix} & 76.75 & 93.30 \\
Manifold\textsubscript{\textcolor{blue}{(ICML'19)}} \cite{verma2019manifold} & 76.85 & 93.50 \\
Mixup\textsubscript{\textcolor{blue}{(ICLR'18)}} \cite{mixup} & 77.03 & 93.52 \\
CutMix\textsubscript{\textcolor{blue}{(ICCV'21)}} \cite{cutmix} & 77.08 & 93.45 \\
Guided-SR\textsubscript{\textcolor{blue}{(AAAI'23)}} \cite{kim2020puzzle} & 77.20 & 93.66 \\
PixMix\textsubscript{\textcolor{blue}{(CVPR'22)}} \cite{hendrycks2022pixmix} & 77.40 & - \\
PuzzleMix\textsubscript{\textcolor{blue}{(ICML'20)}} \cite{kim2020puzzle} & 77.51 & 93.76 \\
GuidedMixup \textsubscript{\textcolor{blue}{(AAAI'23)}} \cite{kang2023guidedmixup} & 77.53 & 93.86 \\
Co-Mixup\textsubscript{\textcolor{blue}{(ICLR'21)}} \cite{kim2020co} & 77.63 & 93.84 \\ 
YOCO\textsubscript{\textcolor{blue}{(ICML'22)}}\cite{han2022yoco} & 77.88 & - \\
Azizi et al.\textsubscript{\textcolor{blue}{(arXiV'23)}} \cite{azizi2023synthetic} & 78.17 & - \\
\hline
\textbf{\textsc{DiffuseMix}} & \textbf{78.64} &  \textbf{95.32}  \\
\hline
\end{tabular}
}
\vspace{-0.9em}
\label{tab:supp_imagenet}
\end{table}
\par
In Table \ref{tab:supp_imagenet}, we provide a comparison of various methods in terms of Top-1 and Top-5 accuracies on ImageNet, specifically when training ResNet-50 as per the training configuration of in \citet{kang2023guidedmixup}. It starts with the baseline Vanilla ResNet model, showing accuracies of \numbersBlue{$75.97\%$} for Top-1 and \numbersBlue{$92.66\%$} for top-5. Various techniques, including Azizi et al., AugMix, Manifold, Mixup, CutMix, Guided-SR, PixMix, PuzzleMix, GuidedMixup, Co-Mixup, YOCO, and another entry from Azizi et al., display a range of improvements, with Top-1 accuracies spanning from \numbersBlue{$69.24\%$} to \numbersBlue{$78.17\%$} and top-5 accuracies (when provided) ranging up to \numbersBlue{$93.86\%$}. The most notable performance is observed in the \textsc{DiffuseMix} method, which outperforms the others by achieving the highest accuracies at \numbersBlue{$78.64\%$}  for top-1 and 
\numbersBlue{$95.32\%$} for top-5. 

\begin{table}[t]
\renewcommand{\arraystretch}{1.0}
\caption{Top-$1$ accuracy comparison on fine-grained visual classification task while training from scratch on ResNet-$50$.}
\label{tab:fgvc_acc_supp}
\vspace{-7pt}
\centering
\scalebox{0.85}{
\begin{tabular}{p{0.17cm}|p{2.7cm}|p{1.2cm}<{\centering}|p{1.2cm}<{\centering}|p{1.2cm}<{\centering}}
\hline\thickhline
\rowcolor{mygray} 
\multicolumn{2}{c|}{}  & \multicolumn{3}{c}{Top-1 Accuracy (\%)} \\ \cline{3-5}
\rowcolor{mygray} 
\multicolumn{2}{c|}{\multirow{-2}{*}{Methods}} & CUB & Aircraft & Cars 
\\ \hline
\multirow{7}{*}{ \rotatebox{90}{ \emph{automated}}} 
& Vanilla ~\cite{resnet} & 65.50 & 80.29 & 85.52 \\ \hline
& Auto Aug \cite{autoaugment} & - & 82.28 & 88.04 \\
& Fast AA \cite{lim2019fast} & - & 82.20 & 87.19 \\
& DADA \cite{li2020dada} & - & 81.16 & 87.14 \\
& RA \cite{cubuk2020randaugment} & - & 82.30 & 87.79 \\
& AdaAug \cite{cheung2021adaaug} & - & 82.50 & 88.49 \\
\hline
\multirow{9}{*}{ \rotatebox{90}{ \emph{mixup family} }} 
& Mixup ~\cite{mixup} & 71.33 & 82.38 & 88.14 \\
& CutMix ~\cite{cutmix} & 72.58 & 82.45 & 89.22 \\
& SaliencyMix ~\cite{uddin2020saliencymix} & 66.66 & 83.14 & 89.04 \\
& Guided-SR ~\cite{kang2023guidedmixup} & 74.08 & 83.51 & 89.23 \\
& SnapMix ~\cite{huang2021snapmix} & 75.53 & 82.96 & 90.10 \\
& PuzzleMix ~\cite{kim2020puzzle} & 74.85 & 82.66 & 89.68 \\
& Co-Mixup ~\cite{kim2020co} & 72.83 & 83.57 & 89.53 \\
& GuidedMixup ~\cite{kang2023guidedmixup} & 77.08  & 
84.32  & 90.27  \\
\hline
& \textbf{\textsc{DiffuseMix}}  & \textbf{79.37} & \textbf{85.76} & \textbf{91.26} \\
\hline
\end{tabular}
}
\end{table}

\subsection{Fine-Grained Visual Classification }
Table \ref{tab:fgvc_acc_supp} presents a comparison of Top-1 accuracy of various methods on a fine-grained visual classification task, using ResNet-50. The methods are categorized into two main groups: automated methods and the mixup family, and are evaluated across three datasets: CUB, Aircraft, and Cars.
\par
In the automated data augmentation, the Vanilla method achieves \numbersBlue{$65.50\%$}, \numbersBlue{$80.29\%$}, and \numbersBlue{$85.52\%$} accuracy on CUB, Aircraft, and Cars respectively. Other automated methods like Auto Aug, Fast AA, DADA, RA, and AdaAug show varied performance, with AdaAug topping this category with accuracies of \numbersBlue{$82.50\%$} for Aircraft and 
\numbersBlue{$88.49\%$} for Cars. The mixup family methods show a notable performance improvement, particularly GuidedMixup demonstrating the accuracies of 
\numbersBlue{$77.08\%$}, \numbersBlue{$84.32\%$}, and \numbersBlue{$90.27\%$} on the three datasets respectively. Nevertheless, our \textsc{ DiffuseMix} stands out by outperforming all compared methods significantly, achieving the highest accuracies of \numbersBlue{$79.37\%$}  for CUB, \numbersBlue{$85.76\%$} for Aircraft, and \numbersBlue{$91.26\%$} for Cars.
\par
This indicates that while both categories of methods enhance performances, mixup family methods demonstrate superior capability in handling fine-grained visual classification tasks. \textsc{DiffuseMix}, in particular, showcases exceptional improvements, suggesting its effectiveness in extracting nuanced features from the images.

\begin{table}
\centering
\caption{Top-1 accuracy on data scarcity experiment using Flower102 dataset where only 10 random images per class are used. Experiments are performed with ResNet-18 network. }
\scalebox{0.95}{
\begin{tabular}{lcc}
\hline\thickhline
\rowcolor{mygray} 
Methods & Valid (\%) & Test (\%) \\
\hline
Vanilla\textsubscript{\textcolor{blue}{(CVPR'16)}}\cite{resnet} & 64.48
 & 59.14  \\ 
Mixup\textsubscript{\textcolor{blue}{(ICLR'18)}}  \cite{mixup} & 70.55  & 66.81 \\
CutMix\textsubscript{\textcolor{blue}{(ICCV'19)}} \cite{cutmix} & 62.68  & 58.51 \\
SaliencyMix\textsubscript{\textcolor{blue}{(ICLR'21)}} \cite{uddin2020saliencymix} & 63.23  & 57.45  \\
Guided-SR\textsubscript{\textcolor{blue}{(AAAI'21)}} \cite{kang2023guidedmixup} & 72.84 & 69.31 \\
SnapMix\textsubscript{\textcolor{blue}{(AAAI'21)}} \cite{huang2021snapmix} &  65.71  & 59.79 \\
PuzzleMix\textsubscript{\textcolor{blue}{(ICML'20)}} \cite{kim2020puzzle} & 71.56  & 66.71 \\
Co-Mixup\textsubscript{\textcolor{blue}{(ICLR'21)}} \cite{kim2020co} & 68.17 & 63.20 \\
GuidedMixup\textsubscript{\textcolor{blue}{(AAAI'23)}} \cite{kang2023guidedmixup}  & 74.74 & 70.44 
\\ 
\hline
 \textbf{\textsc{DiffuseMix}}   & \textbf{77.14}  & \textbf{74.12} \\
\bottomrule
\end{tabular}
}
\label{tab:data_scaricty_supp}
\end{table}

\subsection{Data Scarcity}
Table \ref{tab:data_scaricty_supp} presents the Vanilla method as a baseline with \numbersBlue{$64.48\%$} accuracy on the validation set and \numbersBlue{$59.14\%$} on the test set. SOTA techniques like Mixup and PuzzleMix show improved accuracies, with Mixup achieving \numbersBlue{$70.55\%$} on validation and \numbersBlue{$66.81\%$} on the test set, and PuzzleMix reaching \numbersBlue{$71.56\%$} and \numbersBlue{$66.71\%$}, respectively. 
\par
Notably, the Guided-SR and GuidedMixup methods significantly outperform other approaches, with GuidedMixup achieving the highest accuracies of \numbersBlue{$ 74.74\%$} on validation and \numbersBlue{$70.44\%$} on the test set. Our 
\textsc{DiffuseMix}, which surpasses all compared methods, demonstrates remarkable accuracies of \numbersBlue{$77.14\%$}  on the validation and \numbersBlue{$ 74.12\%$} on the test set, showcasing its superior ability to generalize well from significantly limited data. This evidence suggests that data augmentation and mixing techniques, especially \textsc{DiffuseMix}, are highly beneficial in enhancing model performance under stringent data constraints.
\begin{table}[b]
\centering
\caption{Top-1 (\%) accuracy of self-supervised learning methods. Adding \our{} yields better performance.}
\scalebox{0.90}{
\begin{tabular}{lccc}
\hline
\thickhline
Method & Flower102 & Stanford Cars & Aircraft \\
\hline
\gc{MoCo v2} & \gc{80.31} &  \gc{40.82} &  \gc{51.36} \cr
+ \textsc{DiffuseMix} & \gain{82.15}   & \gain{41.73} & \gain{53.28} \cr
\hline
\gc{SimSiam} & \gc{86.93}  & \gc{48.34} & \gc{40.37} \cr
+ \textsc{DiffuseMix} &  \gain{89.24}  & \gain{49.17} & \gain{42.63} \cr
\hline
\end{tabular}
}
\label{tab:flower_car}
\end{table}

\subsection{Self-Supervised Learning}

Table \ref{tab:flower_car} showcases the Top-1 accuracy of self-supervised learning methods, specifically comparing the performance of MoCo v2 and SimSiam.
\par
\begin{SCtable*}[][t]
\renewcommand{\arraystretch}{1.3}
\setlength{\tabcolsep}{9.6pt}
\caption{On CIFAR-100, \textsc{DiffuseMix} outperforms SOTA on 4 of the 5 distinct safety metrics. Lower is better except for anomaly detection. (SOTA method results are taken from PixMix [17]).}
\vspace{-8pt}
\scalebox{0.85}{
\begin{tabular}{lcccccc}
\hline
\thickhline
 & Mixup & CutMix & AugMix & Outlier & PixMix & \textsc{DiffuseMix} \cr
\hline
Corruptions & 48.0  & 51.5  & 35.4 & 51.5 & 30.5 & \gain{28.5}  \cr
Consistency & 9.5 & 12.0 & 6.5 & 11.3 & 5.7 & \gain{5.1} \cr
Adversaries & 97.4 & 97.0 & 95.6 & 97.2 & 92.9 & \gain{90.2} \cr
Calibration & 13.0 & 29.3 & 18.8 & 15.2 & 8.1 & \gain{7.7} \cr
Anomaly Det. &  71.7 &  74.4 & 84.9 &  \gain{90.3} & 89.3 & 88.3 \cr
\bottomrule
\end{tabular}
}
\label{tab:corruption}
\end{SCtable*}
Initially, MoCo v2 exhibits accuracies of \numbersBlue{$80.31\%$}, \numbersBlue{$40.82\%$}, and \numbersBlue{$51.36\%$} on Flower102, StanfordCars, and Aircraft datasets. After applying \textsc{DiffuseMix} augmentation, it performs better by demonstrating accuracies of \numbersBlue{$82.15\%$}, \numbersBlue{$41.73\%$}, and \numbersBlue{$53.28\%$}. 
SimSiam starts with accuracies of \numbersBlue{$86.93\%$}, \numbersBlue{$48.34\%$}, and \numbersBlue{$40.37\%$}. Adding \textsc{DiffuseMix} as an augmentation method improves the performance to \numbersBlue{$89.24\%$}, \numbersBlue{$49.17\%$}, and \numbersBlue{$42.63\%$}.
This clearly illustrates that integrating \textsc{DiffuseMix} significantly boosts the performance, demonstrating its effectiveness in enhancing self-supervised learning models. 
The systematic gains across different datasets and on multiple methods highlight the robustness of our approach and its potential to improve the accuracies of different machine learning models.

\subsection{Safety Measures}
Table \ref{tab:corruption} showcases a comparative analysis of several data augmentation methods on the CIFAR-100 dataset, focusing on their performance across five different safety metrics. The methods evaluated include Mixup, CutMix, AugMix, Outlier, PixMix, and \textsc{DiffuseMix}. The results highlight \textsc{DiffuseMix}'s superior performance, as it outperforms the state-of-the-art (SOTA) previously established by PixMix in four out of the five categories. \textsc{DiffuseMix}  demonstrates better performance in cases of corruptions, consistency, adversaries, and calibration. In the case of Anomaly detection task, our approach demonstrates comparable performance.

\section{Fractal Dataset}
\label{supp:fractal_dataset}

\begin{figure*}[t]
\centering
\begin{subfigure}{.19\textwidth}
  \centering
  \includegraphics[width=\linewidth]{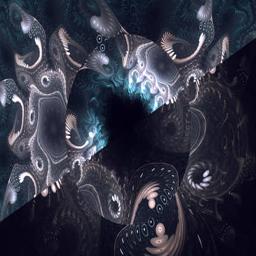}
  \caption{Autumnal Fractal Patterns}
\end{subfigure} \hspace{3pt}
\begin{subfigure}{.19\textwidth}
  \centering
  \includegraphics[width=\linewidth]{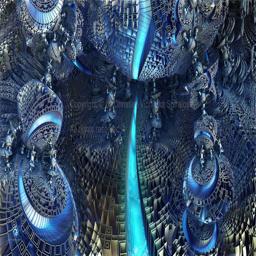}
  \caption{Winter Wonderland}
\end{subfigure}\hspace{4pt}
\begin{subfigure}{.19\textwidth}
  \centering
  \includegraphics[width=\linewidth]{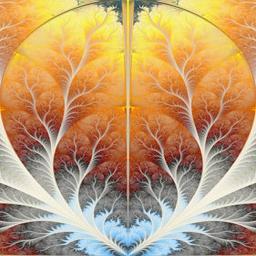}
  \caption{Sunset Hues}
\end{subfigure}\hspace{4pt}%
\begin{subfigure}{.19\textwidth}
  \centering
  \includegraphics[width=\linewidth]{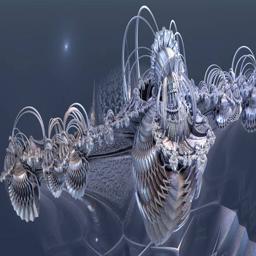}
  \caption{Ukiyo-e Inspired Fractal}
\end{subfigure}\hspace{4pt}%
\begin{subfigure}{.19\textwidth}
  \centering
  \includegraphics[width=\linewidth]{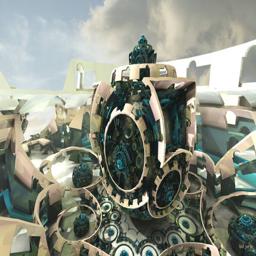}
  \caption{Autumn Reimagined}
\end{subfigure}\hspace{4pt}%

\begin{subfigure}{.19\textwidth}
  \centering
  \includegraphics[width=\linewidth]{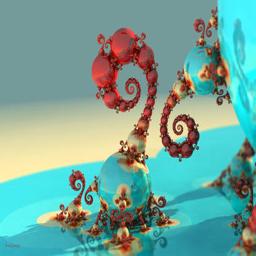}
  \caption{Snowflake Elegance}
\end{subfigure}\hspace{4pt}%
\begin{subfigure}{.19\textwidth}
  \centering
  \includegraphics[width=\linewidth]{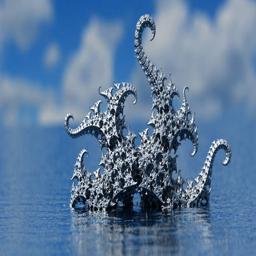}
  \caption{Dusk's Fractal Canvas}
\end{subfigure}\hspace{4pt}%
\begin{subfigure}{.19\textwidth}
  \centering
  \includegraphics[width=\linewidth]{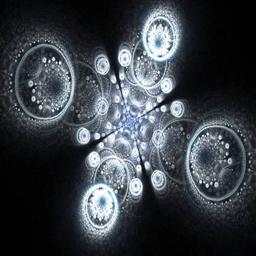}
  \caption{East Meets West}
\end{subfigure}\hspace{4pt}%
\begin{subfigure}{.19\textwidth}
  \centering
  \includegraphics[width=\linewidth]{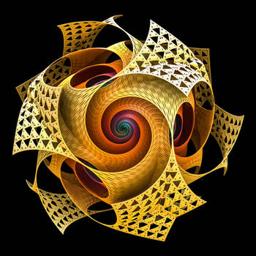}
  \caption{Seasonal Shifts}
\end{subfigure}\hspace{4pt}%
\begin{subfigure}{.19\textwidth}
  \centering
  \includegraphics[width=\linewidth]{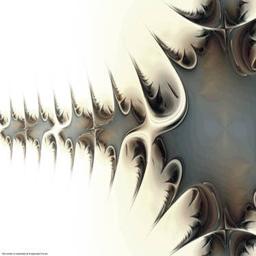}
  \caption{Frozen Fractal Patterns}
\end{subfigure}\hspace{4pt}%
\caption{Some samples taken from our collected fractal dataset. Each subfigure represents a unique fractal image, demonstrating the diversity and complexity inherently present in fractal geometry.}
\label{fig:fractal_dataset}
\end{figure*}

We collected a dataset of $100$ fractal images containing complex patterns and scales. Blending these images to the training images introduces a level of abstraction and complexity not commonly found in regular training images. Some of the example fractal images are provided in Figure \ref{fig:fractal_dataset}. As discussed extensively in the manuscript, fractal blending in \our{} helps the network generalize better by adding \textit{contained} noise or perturbations. The ablation studies reported in our manuscript and supplementary suggest that utilizing fractal blending with the generated images helps stabilizing the training and improves the overall convergence.

\subsection{Fractal with SOTA Methods}

Table \ref{tab:fractal_mixup} presents a performance analysis, particularly focusing on the impact of blending fractals with different augmentation methods using CUB-Birds, Aircraft, Stanford Cars, and Flower102 datasets.
\par
Adding fractal blending to the baseline results in performance improvements on CUB-Birds, Aircraft, and Stanford Cars, but a slight decrease in accuracy on the Flower dataset. The baseline method shows performances of \numbersBlue{$65.50\%$} on CUB-Birds, \numbersBlue{$80.29\%$} on Aircraft, \numbersBlue{$85.52\%$} on Cars and \numbersBlue{$78.34\%$} on Flower102. The Mixup, CutMix, and PuzzleMix methods, when used without fractal, generally show higher accuracy than the baseline, especially on the Stanford Cars and Aircraft datasets. However, the integration of fractal blending with these methods leads to a significant drop in performance across all datasets, suggesting that fractal blending may not be properly aligned with these particular augmentation techniques.
\par
In contrast, when fractals are blended with the hybrid images (Hybrid $H_{iju}$) in our approach, performance improvements are notably observed in three of the four datasets including Aircraft, Stanford Cars, and Flower102. 
datasets, this combination leads to improvements in accuracy, indicating a positive synergy between the hybrid images and fractal blending. However, there's a slight decrease in accuracy for the CUB-Birds dataset.

\begin{table}[t]
\centering
\caption{
Performance comparison (\%) of fractal blending with baseline and other augmentation methods, it is more effective when fractals are blended with our hybrid images \(H_{iju}\).}
\scalebox{0.95}{
\begin{tabular}{lcccc}
\hline\thickhline
Method & CUB-Birds & Aircraft & Cars & Flower \cr
\midrule
\gc{Baseline} & \gc{65.50} &  \gc{80.29} &  \gc{85.52} &  \gc{78.73} \cr
+ \textsc{Fractal} & \gain{66.17}  & \gain{81.27} & \gain{86.73} & \loss{78.34} \cr
\midrule
\gc{Mixup} & \gc{71.33} &  \gc{82.38} &  \gc{88.14} &  \gc{78.12} \cr
+ \textsc{Fractal} & \loss{43.25}  & \loss{44.27} & \loss{54.25} & \loss{57.27} \cr
\midrule
\gc{CutMix} & \gc{72.58}  & \gc{82.45} & \gc{89.22} &  \gc{74.36} \cr
+ \textsc{Fractal} & \loss{46.74} & \loss{41.47} & \loss{56.37} & \loss{52.28} \cr
\midrule
\gc{PuzzleMix} & \gc{74.85}  & \gc{82.66} & \gc{89.68} & \gc{71.68} \cr
+ \textsc{Fractal} & \loss{51.61} & \loss{53.38} & \loss{61.42} & \loss{63.73} \cr
\midrule
\gc{Hybrid (\(H_{iju}\))} & \gc{80.27}  & \gc{85.31} & \gc{ 90.59}  & \gc{79.22} \cr
+ \textsc{Fractal} & \loss{79.37} & \gain{85.76}  & \gain{92.56} & \gain{80.20}  \cr
\bottomrule
\end{tabular}
}
\vspace{-8pt}
\label{tab:fractal_mixup}
\vspace{-8pt}
\end{table}

\end{document}